\documentclass[12pt]{article}
\usepackage{subfigure}
\usepackage{graphicx,harvard}
\usepackage{amsmath}
\usepackage{amssymb,enumerate}
\usepackage{amsthm}
\usepackage{float}
\usepackage{algorithm}
\usepackage{multirow}
\usepackage{subfigure,verbatim}
\usepackage{setspace,color}
\usepackage[left=2.55cm,top=2.55cm,right=2.55cm,bottom=2.55cm]{geometry}
\citationmode{abbr}
\theoremstyle{plain}
\newtheorem{lem}{Lemma}%[section]
\newtheorem{theorem}{Theorem}%[section]
\newtheorem{corollary}[theorem]{Corollary}

\theoremstyle{remark}
\newtheorem{remark}{Remark}

\theoremstyle{definition}
\newtheorem{assumption}{Assumption}

\usepackage[normalem]{ulem}
\usepackage{pdflscape}
\begin{document}
\title{The cluster graphical lasso for improved estimation of Gaussian graphical models}
\author{Kean Ming Tan, Daniela Witten, and Ali Shojaie}
\maketitle

\begin{abstract}
We consider the task of  estimating a Gaussian graphical model in the high-dimensional setting.  The graphical lasso, which involves  maximizing the Gaussian log likelihood subject to an  $\ell_1$ penalty, is a well-studied approach for this task. We begin by introducing a surprising connection between the graphical lasso and hierarchical clustering: the graphical lasso in effect performs a two-step procedure, in which (1) single linkage hierarchical clustering is performed on the variables in order to identify connected components, and then (2) an $\ell_1$-penalized log likelihood is maximized on the subset of variables within each connected component. In other words, the graphical lasso determines the connected components of the estimated network  via single linkage clustering.
Unfortunately, single linkage clustering is known to perform poorly in certain settings.
Therefore, we propose the \emph{cluster graphical lasso}, which involves clustering the features using an alternative to single linkage clustering,  and then performing the graphical lasso on the subset of variables within each cluster.  We  establish  model selection consistency for this technique, and demonstrate its improved performance relative to the graphical lasso in a simulation study, as well as in applications to an equities data set, a university webpage data set, and a gene expression data set.

\end{abstract}
\doublespacing

\section{Introduction}Graphical models have been extensively used in various domains, including modeling of gene regulatory networks and  social interaction networks.
A graph consists of a set of $p$ nodes, corresponding to random variables, as well as a set of edges joining pairs of nodes.
In a conditional independence graph, the absence of an edge between a pair of nodes indicates a pair of variables that are conditionally independent given the rest of the variables in the data set, and the presence of an edge indicates a pair of conditionally dependent nodes.  Hence, graphical models can be used to compactly represent complex joint distributions using a set of local relationships specified by a graph.  Throughout the rest of the text, we will focus on Gaussian graphical models.

Let $\mathbf{X}$ be a $n \times p$ matrix where $n$ is the number of observations and $p$ is the number of features; the rows of $\bf X$ are denoted as ${\bf x}_1, \ldots, {\bf x}_n$.  Assume that $\mathbf{x}_1,\ldots,\mathbf{x}_n \stackrel{\mathrm{iid}}{\sim} N(\mathbf{0},\mathbf{\Sigma})$ where $\mathbf{\Sigma}$ is a $p \times p$ covariance matrix. 
Under this simple model, there is an equivalence between a zero in the inverse covariance matrix and a pair of conditionally independent variables \cite{MKB79}.  More precisely,  $(\mathbf{\Sigma}^{-1})_{jj'} = 0$ for some $j \ne j'$ if and only if the $j$th and $j'$th features are conditionally independent given the other variables.

Let $\mathbf{S}$ denote the empirical covariance matrix of $\mathbf{X}$, defined as $\mathbf{S} = \mathbf{X}^T \mathbf{X}/n$.  A natural way to estimate $\mathbf{\Sigma}^{-1}$ is via maximum likelihood.  This approach involves maximizing
$$\log \det \mathbf{\Theta}- \text{tr}(\mathbf{S} \mathbf{\Theta})$$
with respect to $\mathbf{\Theta}$, where $\mathbf{\Theta}$ is an optimization variable; the solution $\hat{\mathbf{\Theta}}= \mathbf{S}^{-1}$ serves as an estimate for $\mathbf{\Sigma}^{-1}$.  However, in high dimensional settings where $p \gg n$, $\mathbf{S}$ is singular and is not invertible.
Furthermore, even if $\mathbf{S}$ is invertible, $\hat{\mathbf{\Theta}} = \mathbf{S}^{-1}$ typically contains no elements that are exactly equal to zero.  This corresponds to a graph in which the nodes are fully connected to each other; such a graph does not provide useful information.
To overcome these problems, \citeasnoun{YuanLin07} proposed to maximize the penalized log likelihood
\begin{equation}
\label{penlog}
\text{log det } \mathbf{\Theta} - \text{tr}(\mathbf{S} \mathbf{\Theta}) - \lambda \sum_{j\ne j'}|\Theta_{jj'}| 
\end{equation}
with respect to $\mathbf{\Theta}$, penalizing only the off-diagonal elements of $\mathbf{\Theta}$.  A number of algorithms have been proposed to solve  (\ref{penlog}) \citeaffixed{SparseInv,YuanLin07,Rothman08,YuanGlasso08,Scheinberg10}{among others}.  Note that some authors have considered a slight modification to (\ref{penlog}) in which the diagonal elements  of $\mathbf{\Theta}$ are also penalized.  We refer to  the maximizer of  (\ref{penlog}) as the \emph{graphical lasso solution}; it serves as an estimate for ${\bf \Sigma}^{-1}$. When the nonnegative tuning parameter $\lambda$ is sufficiently large, the estimate will be {sparse}, with some elements exactly equal to zero. These zero elements correspond to pairs of variables that are estimated to be conditionally independent.

 \citeasnoun{WittenFriedman11} and \citeasnoun{MazumderHastie11}  presented the following result:

\begin{theorem}
\label{Witten}
The connected components of the graphical lasso solution with tuning parameter $\lambda$ are the same as the connected components of the undirected graph corresponding to the $p \times p$ adjacency matrix ${\bf A}(\lambda)$, defined as
$$A_{jj'}(\lambda) = \begin{cases}
1 &\mathrm{ if  } \; j = j'\\
1_{(|S_{jj'}| > \lambda)}  &\mathrm{ if  }\;  j \neq j'
\end{cases}.$$

\end{theorem}

\noindent 
Here $1_{(|S_{jj'}| > \lambda)}$  is an indicator variable that equals 1 if $|S_{jj'}| > \lambda$, and equals 0 otherwise.
For instance, consider a partition of the features into two disjoint sets, $D_1$ and $D_2$.  Theorem~\ref{Witten} indicates that if $|S_{jj'}| \le \lambda$ for all $j \in D_1$ and all  $j' \in D_2$, then the features in $D_1$ and $D_2$ are in two separated connected components of the graphical lasso solution.  Theorem~\ref{Witten} reveals that solving problem (\ref{penlog}) boils down to two steps:

\begin{enumerate}[1.]
\item Identify the connected components of the undirected graph with adjacency matrix $\bf A$.
\item Perform graphical lasso with parameter $\lambda$ on each connected component separately.
\end{enumerate}

In this paper, we will show that identifying the connected components in the graphical lasso solution -- that is, Step 1 of the two-step procedure described above -- is equivalent to performing \emph{single linkage hierarchical clustering} (SLC) on the basis of a similarity matrix given by the absolute value of the elements of the empirical covariance matrix $\mathbf{S}$.
However, we know that SLC tends to produce \emph{trailing} clusters in which individual features are merged \emph{one at a time}, especially if the data are noisy and the clusters are not clearly separated \cite{ElemStatLearn}. 
In addition, the two steps of the graphical lasso algorithm are based on the same tuning parameter, $\lambda$, which can be suboptimal.
  Motivated by the connection between the graphical lasso solution and single linkage clustering, we therefore propose a new alternative to the graphical lasso.  We will first perform clustering of the variables using an alternative to single linkage clustering, and then perform the graphical lasso on the subset of variables within each cluster.  Our approach  decouples the cutoff for the clustering step from the tuning parameter used for the graphical lasso problem. This results in improved detection of the connected components in high dimensional Gaussian graphical models, leading to more accurate network estimates. Based on this new approach, we also propose a new method for choosing  the tuning parameter for the graphical lasso problem on the subset of variables in  each cluster, which results in consistent identification of the connected components in the graph.

The rest of the paper is organized as follows. In Section \ref{proposal}, we establish a connection between the graphical lasso and single linkage clustering.  In Section \ref{mainidea}, we present our proposal for cluster graphical lasso, a modification of the graphical lasso that involves discovery of the connected components \emph{via} an alternative to SLC.  We prove model selection consistency of our procedure in Section \ref{consistency}.   Simulation results are in Section \ref{simulation}, and  Section \ref{realdata} contains an application of cluster graphical lasso to an equities data set, a webpage data set, and a gene expression data set.  The Discussion is in Section \ref{discussion}.

\section{Graphical lasso and single linkage clustering}
\label{proposal}
We assume that  the columns of $\bf X$ have been standardized to have mean zero and variance one. Let $\tilde{\mathbf{S}}$ denote the $p \times p$ matrix whose elements take the form $\tilde{S}_{jj'}=|{\bf X}_j^T {\bf X}_{j'}|/n = |S_{jj'}|$ where $\mathbf{X}_j$ is the $j$th column of $\mathbf{X}$.

\begin{theorem}
\label{main}
Let $C_1, \ldots, C_K$ denote the clusters that result from performing single linkage hierarchical clustering (SLC) using similarity matrix $ \tilde{\mathbf{S}}$, and cutting the resulting dendrogram at a height of  $0 \le \lambda \le1$.  Let $D_1, \ldots, D_R$ denote the connected components of the graphical lasso solution with tuning parameter $\lambda$.
Then, $K=R$, and there exists a permutation $\pi$ such that $C_k=D_{\pi(k)}$ for $k=1,\ldots,K$.
\end{theorem}
\noindent Theorem~\ref{main}, which is proven in the Appendix, establishes a surprising connection between two seemingly unrelated techniques: the graphical lasso  and SLC.  The connected components of the graphical lasso solution are \emph{identical} to the clusters obtained by performing SLC based on the similarity matrix $\tilde{\mathbf{S}}$.

\begin{figure}[htp]
\centering
\includegraphics[scale=0.5]{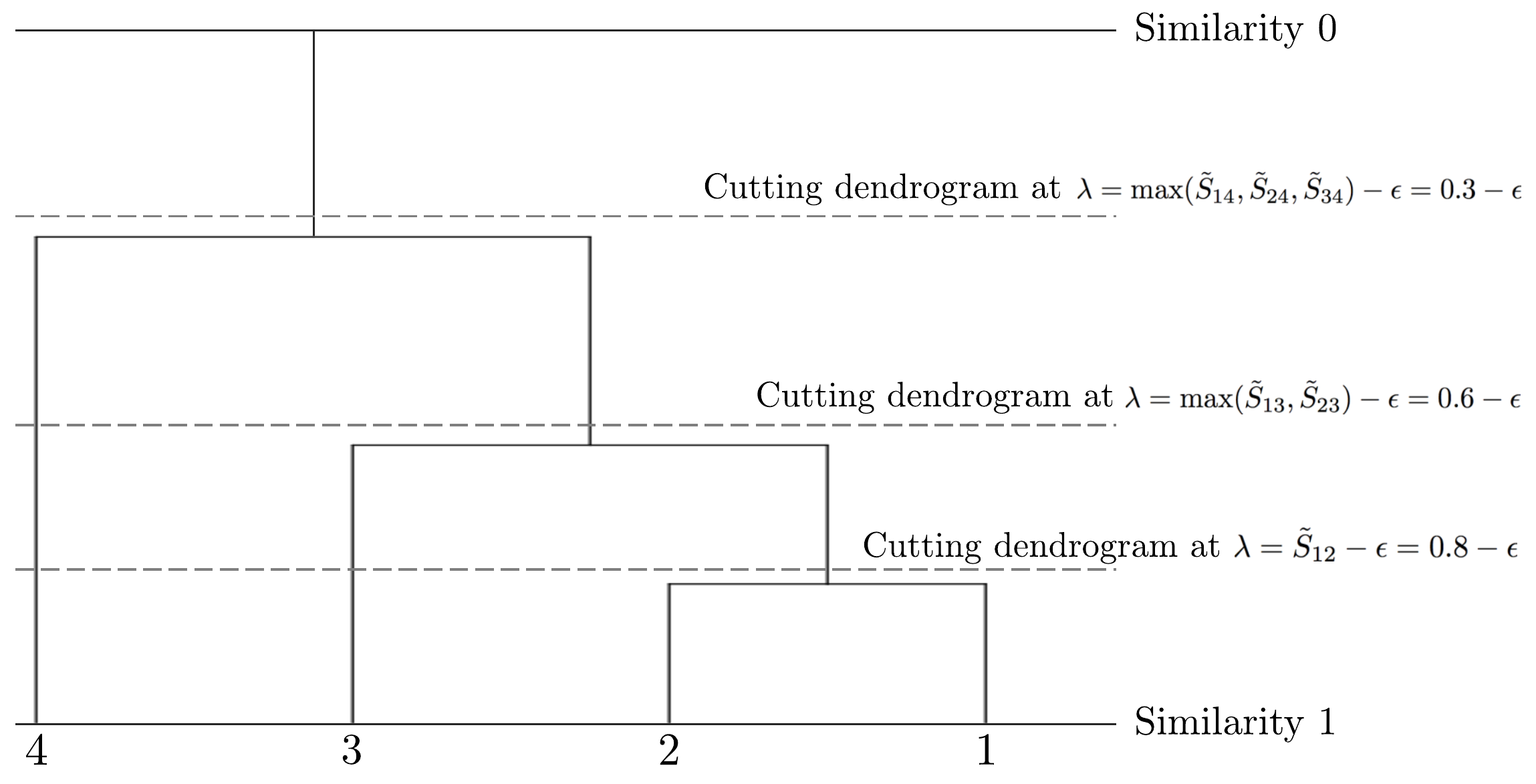}
\caption{SLC is performed on the similarity matrix (\ref{absS}). The resulting dendrogram is shown, as are the clusters that result from cutting the dendrogram at various heights.}
\label{fig:dendrogram}
\end{figure}

Theorem~\ref{main} refers to  \emph{cutting} a dendrogram that results from performing SLC using a similarity matrix. This concept is made clear in Figure~\ref{fig:dendrogram}. In this example,   $\tilde{\bf{S}}$ is given by
\begin{equation} \label{absS}
    \left[
      \begin{array}{cccc}
        1 & 0.8 & 0.6 & 0.3 \\
        0.8 & 1 & 0.5 & 0.2 \\
        0.6 & 0.5 & 1 & 0.1 \\
        0.3 & 0.2 & 0.1 & 1 \\
      \end{array}
    \right].
\end{equation}
For instance, cutting the dendrogram at a height of $\lambda=0.79$ results in three clusters, $\{1,2\}$, $\{3\}$, and $\{4\}$. 
Theorem~\ref{main} further indicates that these are the same as the connected components that result from applying the graphical lasso to $\bf S$ with tuning parameter $\lambda$.

\section{The cluster graphical lasso}
\label{mainidea}

\subsection{A simple alternative to SLC}
Motivated by Theorem~\ref{main}, as well as by the fact that the clusters identified by SLC tend to have an undesirable chain structure \cite{ElemStatLearn}, we now explore an alternative approach, in which we perform clustering before applying the graphical lasso to the set of features within each cluster.

The \emph{cluster graphical lasso} (CGL) is presented in Algorithm \ref{cgl}.
\begin{algorithm}
\caption{Cluster graphical lasso (CGL)}
\label{cgl}
\begin{enumerate}[1.]
\item Let  $C_1, \ldots, C_K$ be the clusters obtained by performing a clustering method of choice based on the similarity matrix $\tilde{\mathbf{S}}$.  The $k$th cluster contains $|C_k|$ features.
\item For $k=1,\ldots,K$:
\begin{enumerate}[(a)]
\item  Let $\mathbf{S}_k$
be the empirical covariance matrix for the features in the $k$th cluster.  Here, $\mathbf{S}_k$ is a $|C_k| \times |C_k|$ matrix. 
\item Solve the graphical lasso problem (\ref{penlog}) using the covariance matrix  $\mathbf{S}_k$ with a given value of $\lambda_k$.  Let $\hat{\mathbf{\Theta}}_k$ denote the graphical lasso estimate.
\end{enumerate}
\item Combine the $K$ resulting graphical lasso estimates into a $p \times p$ matrix $\hat{\mathbf{\Theta}}$ that is block diagonal with blocks $\hat{\mathbf{\Theta}}_1, \ldots, \hat{\mathbf{\Theta}}_K$.
\end{enumerate}
\end{algorithm}
We partition the features into $K$ clusters  based on $\tilde{\mathbf{S}}$, and then perform the graphical lasso estimation procedure on the subset of variables within each cluster.
Provided that the true network has several connected components, and that the clustering technique that we use in Step 1 is better than SLC, we expect CGL to outperform graphical lasso. 

Furthermore, we note that by Theorem~\ref{main}, the usual graphical lasso is a special case of Algorithm~\ref{cgl}, in which the clusters in Step 1 are obtained by cutting the SLC dendrogram at a height $\lambda$,
 and in which $\lambda_1=\ldots=\lambda_K=\lambda$ in Step 2(a). 

The advantage of CGL over the graphical lasso is two-fold.
\begin{enumerate}[1.]
\item As mentioned earlier, SLC often performs poorly, often resulting in estimated graphs with one large connected component, and many very small ones. Therefore, identifying the connected components  using a better clustering procedure may yield improved results.
\item As revealed by Theorems~\ref{Witten} and \ref{main}, the graphical lasso effectively couples two operations using a single tuning parameter $\lambda$: identification of the connected components in the network estimate, and identification of the edge set within each connected component. Therefore, in order for the graphical lasso to yield a solution with many connected components, each connected component must typically be extremely sparse. CGL allows for these two operations to be decoupled, often to advantage.
\end{enumerate}

\subsection{Interpretation of CGL as a penalized log likelihood problem}
Consider the optimization problem
\begin{equation}
\underset{\mathbf{\Theta}} {\text{minimize}} \{ -\log \det \mathbf{\Theta}+ \text{tr}(\mathbf{S\Theta}) + \sum_{ j\ne j'} w_{jj'} |\Theta_{jj'}| \},
\label{penCGL}
\end{equation}
where $$w_{jj'} = \begin{cases} \lambda_k& \text{if } j,j'\in C_k \\ \infty & \text{if } j \in C_k, j' \in C_{k'}, k \neq k' \end{cases}.$$
By inspection, the solution to this problem is the CGL network estimate.   In other words, the CGL procedure amounts to  solving a penalized log likelihood problem in which we impose an arbitrarily large penalty on $|\Theta_{jj'}|$ if the $j$th and $j'$th features are in different clusters.  % Then, the solution to (\ref{penCGL}) is the same as the CGL estimate.
In contrast, if $w_{jj'}=\lambda$ in (\ref{penCGL}), then this amounts to the graphical lasso optimization problem (\ref{penlog}).  % we get from this optimization problem will be exactly the same as that of CGL.

\subsection{Tuning parameter selection}
CGL involves several tuning parameters: the number of clusters $K$ and the sparsity parameters $\lambda_1,\ldots,\lambda_K$.  It is well-known that selecting tuning parameters in unsupervised settings is a challenging problem \citeaffixed{MC85,Gordon96,TWH2000,ElemStatLearn,StabilitySelection}{for an overview and several past proposals, see e.g.}.
Algorithm~\ref{selectK} outlines an approach for selecting $K$.  It involves leaving out random elements from the matrix $\tilde{\bf S}$ and performing clustering. The clusters obtained are then used to impute the left-out elements, and the corresponding mean squared error is computed.  Roughly speaking, the optimal $K$ is that for which the mean squared error is smallest.  This is related to past approaches in the literature for performing
 tuning parameter selection in the unsupervised setting by recasting the unsupervised problem as a supervised one \citeaffixed{Wold1978,OwenPerry2008,WittenHastieTibs09}{see e.g.}.
 The numerical investigation in Section~\ref{simulation} indicates that our algorithm results in reasonable estimates of the number
 of connected components, and that the performance of CGL is not very sensitive to the value of $K$.

In Corollary \ref{chooselambdacorr}, we propose a choice of $\lambda_1,\ldots,\lambda_K$ that guarantees consistent recovery of the connected components.

\begin{algorithm}
\caption{Selecting the number of clusters $K$}
\label{selectK}
\begin{enumerate}[1.]
\item Repeat the following procedure $T$ times:

\begin{enumerate}[(a)]
\item Let $\mathcal{M}$ be a set that contains ${p (p-1)}/{2T}$ elements of the form $(i,j)$,  where $(i,j)$ is drawn randomly from $\{(i,j): i,j\in \{1,\ldots,p \}, i < j  \}$.  Augment the set $\mathcal{M}$ such that if $(i,j) \in \mathcal{M}$, then $(j,i) \in \mathcal{M}$.  We refer to $\mathcal{M}$ as a set of \emph{missing} elements.

\item Construct a $p \times p$ matrix, $\tilde{\mathbf{S}}^*$, for which the elements in $\mathcal{M}$ are removed and are replaced by taking the average of the corresponding row and column means of the non-missing elements in $\tilde{\mathbf{S}}$:

\begin{equation}
\tilde{S}^*_{ij} = \begin{cases} \tilde{S}_{ij} & \mbox{ if } (i,j) \notin \mathcal{M}\\
\displaystyle 0.5\left(\sum_{j' \in \mathcal{M}^c_{i}} \tilde{S}_{ij'} / |\mathcal{M}^c_{i}| + \displaystyle \sum_{i'\in \mathcal{M}^c_{j}} \tilde{S}_{i'j} /|\mathcal{M}^c_{j}|\right) & \mbox{ if } (i,j)  \in \mathcal{M}
\end{cases},
\end{equation}
where $\mathcal{M}^c_{i} = \{ j : (i,j)\notin \mathcal{M}  \}$ and $|\mathcal{M}^c_i|$ is the cardinality of $\mathcal{M}^c_i$. 
\item For each value of $K$ under consideration:

\begin{enumerate}[i.]
\item Perform the clustering method of choice based on the similarity matrix $\tilde{\mathbf{S}}^*$.  Let $C_1, \ldots, C_K$ denote the clusters obtained.

\item Construct a $p \times p$ matrix $\mathbf{B}$ in which each element is imputed using the block structure of $\tilde{\mathbf{S}}^*$ based on the clusters $C_1,\ldots,C_K$ obtained:

\begin{equation}
{B}_{ij} = \begin{cases} \displaystyle \frac{\sum_{i',j' \in C_k, i'\ne j'} \tilde{S}^*_{i'j'}}{|C_k|\times (|C_k|-1)} & \mbox{ if } i,j  \in C_k \mbox{ and } i\ne j\\
\displaystyle \frac{\sum_{k \ne k'} \sum_{i'\in C_k, j'\in C_{k'}} \tilde{S}^*_{i'j'}}{p^2-\sum_{k} |C_k|^2} & \mbox{ if } i\in C_k, j\in C_{k'}, \mbox{and } k \ne k' \\
\end{cases},
\end{equation}

\item Calculate the mean squared error as follows:
\begin{equation}
\label{MSEK}
\displaystyle{\sum_{(i,j)\in \mathcal{M}}} (\tilde{S}_{ij}-B_{ij})^2/|\mathcal{M}|.
\end{equation}

\end{enumerate}
\end{enumerate}
\item For each value of $K$ that was considered in Step 1(c), calculate $m_K$, the mean of quantity (\ref{MSEK}) over the iterations, as well as $s_K$, its standard error.
\item Identify the set $\{ K: m_K \le m_{K+1}+ 1.5 \times s_{K+1} \}$. Select the smallest value in this set.
\end{enumerate}
\end{algorithm}

% $\forall i,j$ in the same cluster and set $w_{ij}$ to be very large if feature $i,j$ are in different clusters.  This is analogous to the optimization problem of CGL.

\section{Consistency of cluster graphical lasso}\label{consistency}
In this section, we establish that CGL consistently recovers the connected components of the underlying graph, as well as its edge set.
%Theoretical properties of the graphical lasso problem (\ref{penlog}) has been studied in \citeaffixed{Rothman08}{see e.g.}.
A number of authors have shown consistency of the graphical lasso solution for different matrix norms 
\cite{Rothman08,LamFan2009,CaietalJASA11}.  \citeasnoun{LamFan2009} further showed that under certain conditions, the graphical lasso solution is \textit{sparsistent}, i.e., zero entries of the inverse covariance matrix are correctly estimated with probability tending to one.
%They obtained results on the rates of convergence for consistency and sparsistency under Frobenius norm \citeaffixed{LamFan2009,CaietalJASA11}{see e.g.}.
\citeasnoun{LamFan2009} also showed that there is no choice of $\lambda$ that can simultaneously achieve the optimal rate of sparsistency and consistency  for estimating $\mathbf{\Sigma}^{-1}$, unless the number of non-zero elements in the off-diagonal entries is no larger than $O(p)$.
%\textcolor{red}{maybe we can say something of our approach, we decoupled the two steps and hence, might get partial sparsistency but consistent estimate of each of the connected components?}
In a more recent work, \citeasnoun{Ravikumar2011} studied the  graphical lasso estimator under a variety of tail conditions, and established that the procedure correctly identifies the structure of the graph, if an incoherence assumption holds on the Hessian of the inverse covariance matrix, and if the minimum non-zero entry of the inverse covariance matrix is sufficiently large. We will restate these conditions more precisely in Theorem~\ref{thm:CGLconsistency}.

Here, we focus on model selection consistency of CGL, in the setting where the inverse covariance matrix is block diagonal.
%Our results parallel the results established in \citet{Ravikumar2011}.
To establish the model selection consistency of CGL, we need to show that (i) CGL correctly identifies the connected components of the graph, and (ii) it correctly identifies the set of edges (i.e. the set of non-zero values of the inverse covariance matrix) within each of the connected components. More specifically, we first show that CGL with clusters obtained from performing  SLC, \emph{average linkage hierarchical clustering} (ALC), or \emph{complete linkage hierarchical clustering} (CLC) based on $\tilde{\mathbf{S}}$ consistently identifies the connected components of the graph. Next, we adapt the results of \citeasnoun{Ravikumar2011} on model selection consistency of graphical lasso in order to establish the rates of convergence of the CGL estimate.

As we will show below, our results highlight the potential advantages of CGL in the settings where the underlying inverse covariance matrix is block diagonal (i.e. the graph consists of multiple connected components).
As a byproduct, we also address the problem of determining the appropriate set of  tuning parameters for penalized estimation of the inverse covariance matrix in high dimensions: given  knowledge of $K$, the number of connected components in the graph, we suggest a choice of $\lambda_1, \ldots, \lambda_K$ for CGL that leads to consistent identification of the connected components in the underlying network.
In the context of the graphical lasso, \citeasnoun{Banerjee} have suggested a choice of $\lambda$ such  that the probability of adding edges between two disconnected components is bounded by $\alpha$, given by
\begin{equation}
\lambda = (\max_{i<j} S_{ii} S_{jj})\frac{t_{n-2}(\alpha/2p^2)}{\sqrt{n-2+t^2_{n-2} (\alpha/2p^2)}},
\label{banerjeelambda}
\end{equation} 
where $t_{n-2}(\alpha)$ denotes the $(100-\alpha)$ percentile of the Student's t-distribution with $n-2$ degrees of freedom, and $S_{ii}$ is the empirical variance of the $i$th variable.
The proposal of \citeasnoun{Banerjee} is based on an earlier result by \citeasnoun{MB2006}, who suggested a similar choice of $\lambda$ for estimating the edge set of the graph using the \textit{neighborhood selection} approach.
Note that (\ref{banerjeelambda})  is fundamentally different from our proposal, as this choice of $\lambda$ does not guarantee that each  connected component is not broken into several distinct connected components.  In fact, empirical studies have found that the choice of $\lambda$ in (\ref{banerjeelambda}) may result in an estimated graph that is too sparse \cite{Shojaieetal2012}.

Before we continue, we summarize some notation that will be used in Sections \ref{consistentrecovery} and \ref{modelselection}.  Let $\mathbf{X}$ be a $n \times p$ matrix; the rows of $\mathbf{X}$ are denoted as $\mathbf{x}_1,\ldots,\mathbf{x}_n$, where $\mathbf{x}_1,\ldots,\mathbf{x}_n \stackrel{\mathrm{iid}}{\sim} N(0,\mathbf{\Sigma})$ and $\mathbf{\Sigma}$ is a block diagonal covariance matrix with $K$ blocks.  We let $C_k$ be the feature set corresponding to the $k$th block.  (In previous sections, $C_1,\ldots,C_K$ denoted a set of estimated clusters; in this section only, $C_1,\ldots,C_K$ are the \emph{true} and in practice unknown clusters.)
Also, let $\hat{C}_1, \ldots, \hat{C}_K$ denote a set of estimated clusters obtained from performing SLC, ALC, or CLC.
In what follows, we use the terms \emph{clusters} and \emph{connected components} interchangeably. Let $\tilde{\mathbf{S}} = |\mathbf{X}^T\mathbf{X}/n|$ be the absolute empirical covariance matrix. Proofs of what follows are provided in the Appendix.

\subsection{Consistent recovery of the connected components}
\label{consistentrecovery}
%We present several lemmas and a corollary in this section.
We now present some results on the recovery of the connected components of $\mathbf{\Sigma}^{-1}$ by SLC, ALC, or CLC, as well as its implications for the CGL procedure.

\begin{comment}
\begin{lem}
\cite{Ravikumar2011} Let $\mathbf{x}_i$ be $i.i.d.$ $N(0,\mathbf{\Sigma})$ and assume that $\Sigma_{ii} \le M$ $\forall i$, where $M$ is some constant.  Given $n$ $i.i.d.$ samples, the associated empirical covariance matrix $\mathbf{S}= \mathbf{X}^T\mathbf{X}/n$ satisfies
\begin{equation*}
P[ |S_{ij} - \Sigma_{ij} | > \delta] \le c_1 \exp(-c_2 n \delta^2),
\end{equation*}
where $c_1$, $c_2$ and $\delta$ are constant depending on $M$ only.
\label{ravikumarresult}
\end{lem}
\end{comment}

\begin{lem}
Assume that $\mathbf{\Sigma}$ is a block diagonal matrix with $K$ blocks and  diagonal elements  $\Sigma_{ii} \le M $ $\forall i$, where $M$ is some constant.     Furthermore, let 
%$ a = \min_{i,j\in C_k : k=1,\ldots, K} |\Sigma_{ij}|$, and $a \ge 2t \sqrt{\frac{\log p}{n}}$ 
$$  \min_{i,j\in C_k : k=1,\ldots, K} |\Sigma_{ij}|  \ge 2t \sqrt{\frac{\log p}{n}}$$
for some $t>0$ such that $c_2 t^2>2$.  Then performing  SLC, ALC, or CLC with similarity matrix $\tilde{\mathbf{S}}$ satisfies $P( \exists k : \hat{C}_k \ne C_k) \le c_1\,p^{2 - c_2t^2}$.
%$P( \exists k : \hat{C}_k \ne C_k) \le \frac{c_1}{p^{c_2t^2-2}}$.
\label{clusterconsistent}
\end{lem}

\noindent Lemma~\ref{clusterconsistent} establishes the consistency of identification of connected components by performing hierarchical clustering using SLC, ALC, or CLC, provided that  $n= \Omega(\log p)$ as $n, p \rightarrow \infty$, and provided that no within-block element of $\bf \Sigma$ is too small in absolute value. \citeasnoun{BuhlmannCluster2012} also commented on the consistency of hierarchical clustering. % on the absolute empirical covariance matrix.

Let $\tilde{\mathbf{S}}_1,\ldots, \tilde{\mathbf{S}}_K$ denote the $K$ blocks of $\tilde{\mathbf{S}}$ corresponding to the features in $\hat{C}_1, \ldots, \hat{C}_K$.  In other words, $\tilde{\mathbf{S}}_k$ is a $|\hat{C}_k| \times |\hat{C}_k|$ matrix.
The following corollary on selecting the tuning parameter $\lambda_1,\ldots,\lambda_K$ for  CGL is a direct consequence of Lemma \ref{clusterconsistent} and Theorem \ref{Witten}.

\begin{corollary}
Assume that the diagonal elements of $\mathbf{\Sigma}$ are bounded and that
$$\min_{i,j\in C_k : k=1,\ldots, K} |\Sigma_{ij}| = \Omega \left( \sqrt{\frac{\log p}{n}} \right).$$
%the minimal element of $\mathbf{\Sigma}_k$ is $\Omega(\sqrt{\frac{\log p}{n}})$ for $k=1,\ldots, K$.
Let $\bar{\lambda}_{k}$ be the smallest value that cuts the dendrogram resulting from applying SLC to $\tilde{\mathbf{S}}_k$ into two clusters.  Performing CGL with SLC, ALC, or CLC and penalty parameter $\lambda_k \in [0,\bar{\lambda}_k)$ for $k=1,\ldots,K$ leads to consistent identification of the $K$ connected components if $n= \Omega(\log p)$ as $n, p \rightarrow \infty$.%at a rate of convergence $O_P (\sqrt{\frac{\log p}{n}})$.
\label{chooselambdacorr}
\end{corollary}

\noindent Corollary \ref{chooselambdacorr}  implies that one can consistently recover the $K$ connected components by (a) performing hierarchical clustering based on $\tilde{\mathbf{S}}$ to obtain $K$ clusters and (b) choosing the tuning parameter $\lambda_k \in [0,\bar{\lambda}_k)$  in Step 2(b) of Algorithm \ref{cgl} for each of the $K$ clusters.
However, Corollary~\ref{chooselambdacorr} does not guarantee that this set of tuning parameters $\lambda_1,\ldots,\lambda_K$ will identify the correct edge set within each connected component. We now establish such a result.
%\end{comment}

%Next, we present  a result to establish model selection consistency of CGL.

\subsection{Model selection consistency of CGL}
\label{modelselection}
The following theorem combines Lemma~\ref{clusterconsistent} with results on model selection consistency of the graphical lasso  \cite{Ravikumar2011} in order to establish the model selection consistency of CGL. We start by introducing some notation and stating the assumptions needed.

%Recall that the rows of $\mathbf{X}$, $\mathbf{x}_1,\ldots, \mathbf{x}_n \overset{iid}\sim N(0,\mathbf{\Sigma})$, where $\mathbf{\Sigma}$ is a $p\times p$ matrix.
Let $\mathbf{\Theta} = \mathbf{\Sigma}^{-1}$. $\bf \Theta$ specifies an  undirected graph with $K$ connected components; the $k$th connected component has edge set $E_k = \{(i,j)\in C_k : i \ne j, \theta_{ij} \ne 0  \}$.  Let $E= \cup_{k=1}^K E_k$.  Also, let $F_k = E_k \cup \{(j,j): j\in C_k \}$; this is  the union of the edge set and the diagonal elements for the $k$th connected component.  Define $d_k$ to be the maximum degree in the $k$th connected component, and let $d= \max_{k} d_k$.  Also, define $p_k = |C_k|$, $p_{\min} =\min_{k} p_k $, and $p_{\max} =\max_{k} p_k $.

Assumption \ref{A1} involves the Hessian of Equation \ref{penlog}, which takes the form
\[
\mathbf{\Gamma}^{(k)} = \nabla^2_{\mathbf{\Theta}_k} \{-\log \det (\mathbf{\Theta}_k) \} = \mathbf{\Sigma}_k \otimes \mathbf{\Sigma}_k,
\]
where $\otimes$ is the Kronecker matrix product, and  $\mathbf{\Gamma}^{(k)}$ is a $p^2_k \times p^2_k$ matrix. With some abuse of notation, we define ${\bf\Gamma}^{(k)}_{AB}$ as the $|A| \times |B|$ submatrix  of ${\bf\Gamma}^{(k)}$ whose rows and columns are indexed by $A$ and $B$ respectively, i.e., ${\bf \Gamma}^{(k)}_{AB} = [\mathbf{\Sigma}_k \otimes \mathbf{\Sigma}_k]_{AB} \in \mathbb{R}^{|A|\times |B|}$. %\cite{Ravikumar2011}. %and that each element of $\mathbf{\Gamma}$ can be written as $\Gamma_{(j,k),(l,m)} = \text{Cov}(X_jX_k,X_l X_m)$, where $X_j$ is the $j$th variable  \cite{Ravikumar2011}.

\begin{assumption}
%\cite{Ravikumar2011}
There exists some $\alpha \in (0,1]$ such that for all $k=1,\ldots,K$,
\[
\max_{e\in \{ (C_k \times C_k)\setminus F_k\} } \| \mathbf{\Gamma}^{(k)}_{eF_k} (\mathbf{\Gamma}^{(k)}_{F_kF_k})^{-1}    \|_1 \le (1-\alpha).
\]
%where $\| . \|$ is the $\ell_1$ norm.
\label{A1}
\end{assumption}

\begin{assumption}
%\cite{Ravikumar2011}
For $\theta_{\min}  := \min_{(i,j)\in E} |\theta_{ij}|$, the minimum non-zero off-diagonal element of the inverse covariance matrix,
%\[
%\theta_{\min} := \min_{(i,j)\in E} |\Theta_{ij}|  = \Omega \left( \sqrt{\frac{\log p}{n}} \right),
%\]
 $$\theta_{\min} = \Omega \left( \sqrt{\frac{\log p}{n}} \right)$$ as $n$ and $p$ grow.
\label{A2}
\end{assumption}

We now present our main theorem.  It relies heavily on Theorem 2 of \citeasnoun{Ravikumar2011}, to which we refer the reader for details.
\begin{theorem}
Assume that $\mathbf{\Sigma}$  satisfies the conditions in Lemma~\ref{clusterconsistent}. Further, assume that Assumptions \ref{A1} and \ref{A2} are satisfied, and that $\|{\bf \Sigma}_k \|_{\infty}$ and $\| ({\bf \Gamma}^{(k)}_{F_k,F_k} )^{-1} \|_{\infty}$ are bounded,
where $\| {\bf A} \|_{\infty}$ denotes the $\ell_{\infty}$ norm of $\bf A$.
Assume $K = O(p_{\min})$,    $\tau > 3$, where $\tau$ is a user-defined parameter,
%$\alpha \in (0,1]$ for which Assumption \ref{A1} is satisfied,
and
\[
    n = \Omega\left( \log(p) \vee  d^2\log(p_{\max}) \right) \mathrm{\; as \;} n, p_{\min}\rightarrow \infty.
\]
Let $\hat{E}$ denote the edge set from the CGL estimate using SLC, ALC, or CLC with $K$ clusters and $\lambda_k = \frac{8}{\alpha} \sqrt{\frac{c_2 (\tau \log p_k + \log 4)}{n}}$.
%Then $\hat{E} = E$ with probability at least $1 - \frac{c_1}{p^{c_2t^2-2}} - \frac{K}{p_{\min}^{\tau - 2}}$.
Then $\hat{E} = E$ with probability at least $1 - c_1\,p^{2-c_2t^2} - K\,p_{\min}^{2 - \tau}$.
\label{thm:CGLconsistency}
\end{theorem}

\begin{remark}
Theorem \ref{thm:CGLconsistency} states that CGL with SLC can improve upon existing model selection consistency results for the graphical lasso. Recall that Theorem~\ref{main} indicates that graphical lasso is a two-step procedure in which SLC precedes precision matrix estimation within each cluster. However, the two steps in the graphical lasso procedure involve a single tuning parameter, $\lambda$.  The improved rates for CGL in Theorem 4 are achieved by decoupling the choice of tuning parameters in the two steps. 
\end{remark}

\begin{remark}
Note that the assumption of Lemma~\ref{clusterconsistent} that  $\Sigma_{ij} \neq  0$ for all $(i,j) \in C_k$ does not require the underlying conditional independence graph for that connected component to be fully connected. For example, consider the case where connected components of the underlying graph are \textit{forests}, with non-zero partial correlations on each edge of the graph. Note  that there exists a unique path $\mathcal{P}_{ij}$ of length $q$ between each pair of elements $i$ and $j$ in the same block. Then by Theorem 1 of \citeasnoun{jones2005covariance},
\[
    \Sigma_{ij} = c \, \theta_{ik_1} \theta_{k_1k_2} \ldots \theta_{k_{q-1}j},
\]
for some constant $c$ depending on the path $\mathcal{P}_{ij}$ and the determinant of the precision matrix. By the positive definiteness of precision matrix, and the fact that partial correlations along the edges of the graph are non-zero, it follows that $\Sigma_{ij} \ne 0$ for all $i$ and $j$ in the same block.
\end{remark}

\begin{remark}
The convergence rates in Theorem~\ref{thm:CGLconsistency} can result in improvements over existing rates for the graphical lasso estimator. For instance, suppose the graph consists of $K = m^a$ connected components of size $p_k = m^b$ each, for positive integers $m$, $a$, and $b$ such that $m>1$ and $a<b$. Also, assume that $d = m$. Then, based on the results of \citeasnoun{Ravikumar2011}, consistency of the graphical lasso requires $n = \Omega(m^2 (a+b) \log(m)  )$ samples, whereas Theorem~\ref{thm:CGLconsistency} implies that $n = \Omega(b m^2 \log(m) )$ samples suffice for consistent estimation using CGL.
\label{convergenceremark}
\end{remark}

Remark \ref{convergenceremark} is not surprising.  The reduction in the required sample size for CGL is achieved from the extra information on the number of connected components in the graph.  This result suggests that decoupling the identification of connected components and estimation of the edges  can result in improved estimation of Gaussian graphical models in the settings where $\mathbf{\Sigma}^{-1}$ is block diagonal.  The results in the next two sections provide empirical evidence in support of these findings.

\section{Simulation study}
\label{simulation}
We consider two simulation settings: $\mathbf{\Sigma}^{-1}$ is block diagonal with two blocks, and $\mathbf{\Sigma}^{-1}$ is approximately block diagonal with two blocks.

\begin{algorithm}
\caption{Generation of   sparse inverse covariance matrix $\mathbf{\Sigma}^{-1}$}
\label{precisionmatrix}
\begin{enumerate}
\item  Let $C_1,\ldots, C_K$ denote a partition of the $p$ features. % into $K$ connected components.

\item For $k=1,\ldots,K$, construct a $|C_k| \times |C_k|$ matrix $\mathbf{A}_k$ as follows.  For each $j < j'$,

%each $\mathbf{\Sigma}^{-1}_{C_r}$:

\begin{equation}
({\bf A}_k)_{jj'} =
\begin{cases} \mbox{Unif}(0.25, 0.75) & \mbox{ with probability }  (1-s) \times 0.75  \\
 \mbox{Unif}(-0.75, -0.25) & \mbox{ with probability } (1-s) \times 0.25  \\
0 & \mbox{ with probability } s
\end{cases},
\end{equation}
where $s$ is the level of sparsity in $\mathbf{A}_k$.  Then, we set $(\mathbf{A}_k)_{j'j} = (\mathbf{A}_k)_{jj'}$ to obtain symmetry.  Furthermore, set the diagonal entries of ${\bf A}_k$ to equal zero.
\item Create a matrix ${\bf \Sigma}^{-1}$ that is block diagonal with blocks $\mathbf{A}_1, \ldots, \mathbf{A}_K$.% and set the diagonal entries of ${\bf \Sigma}^{-1}$ to equal zero.

\item In order to achieve positive definiteness of $\mathbf{\Sigma}^{-1}$, calculate the minimum eigenvalue $e_{\min}$ of $\mathbf{\Sigma}^{-1}$. For $j=1,\ldots,p$,  set
\begin{equation}
({\bf \Sigma}^{-1})_{jj} =
%\begin{cases}  (\mathbf{\Sigma}^{-1})_{jj} +  0.1 & \mbox{   if  } e_{min} \ge 0  \\
 - e_{\min}+0.1  \qquad \mbox{ if } e_{\min} < 0.
% \end{cases}
 \end{equation}
% \noindent for all $j= 1, \ldots, p$.
\end{enumerate}
\end{algorithm}

\subsection{  ${\bf \Sigma}^{-1}$ block diagonal}
\label{simulation2}

We generated a $n\times p$ data matrix $\mathbf{X}$ according to $\mathbf{x}_{1},\ldots,\mathbf{x}_n \stackrel{\mathrm{iid}}{\sim}  N(\mathbf{0},\mathbf{\Sigma})$, where $\mathbf{\Sigma}^{-1}$ is a block diagonal inverse covariance matrix with two equally-sized blocks, and sparsity level $s=0.4$ within each block, generated using Algorithm~\ref{precisionmatrix}.
%We use the procedure in Algorithm~\ref{precisionmatrix} to generate $\mathbf{\Sigma}^{-1}$.
We first standardized the variables to have mean zero and variance one.  Then, we performed the graphical lasso as well as CGL using ALC with (a) the tuning parameter $K$ selected using Algorithm \ref{selectK} and (b) $K = \{2, 4 \}$.  For simplicity, we chose $\lambda_1 = \ldots = \lambda_K = \lambda$ where $\lambda$ ranges from $0.01$ to $1$.  We considered two cases in our simulation: $n=50, p=100$  and $n=200, p=100$.  Results are presented in Figure~\ref{n50p100k2}.  More extensive simulation results are presented in the Supplementary Materials.

\begin{figure}[htp]
\begin{center}
\includegraphics[scale=0.45]{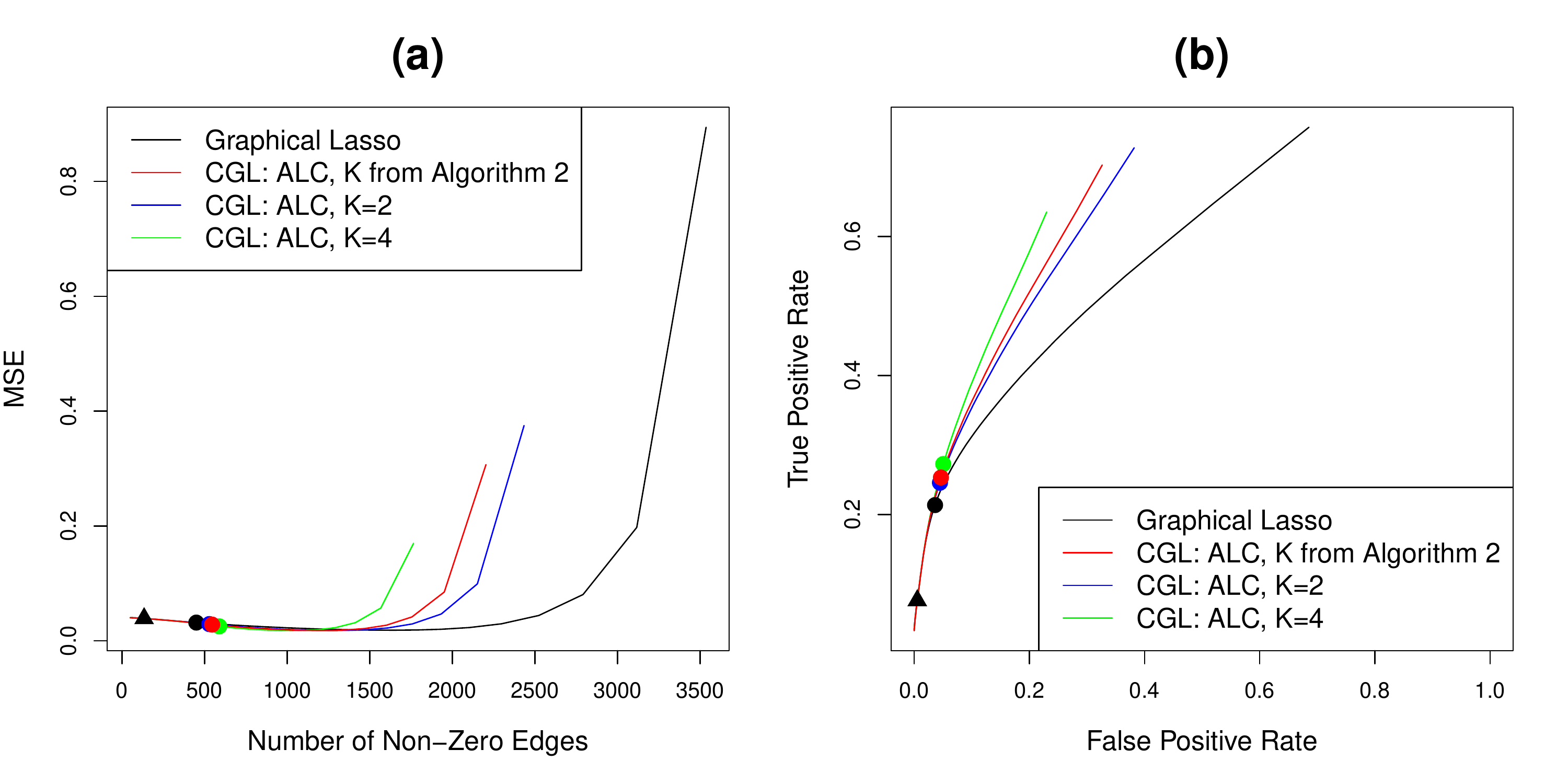}
\includegraphics[scale=0.45]{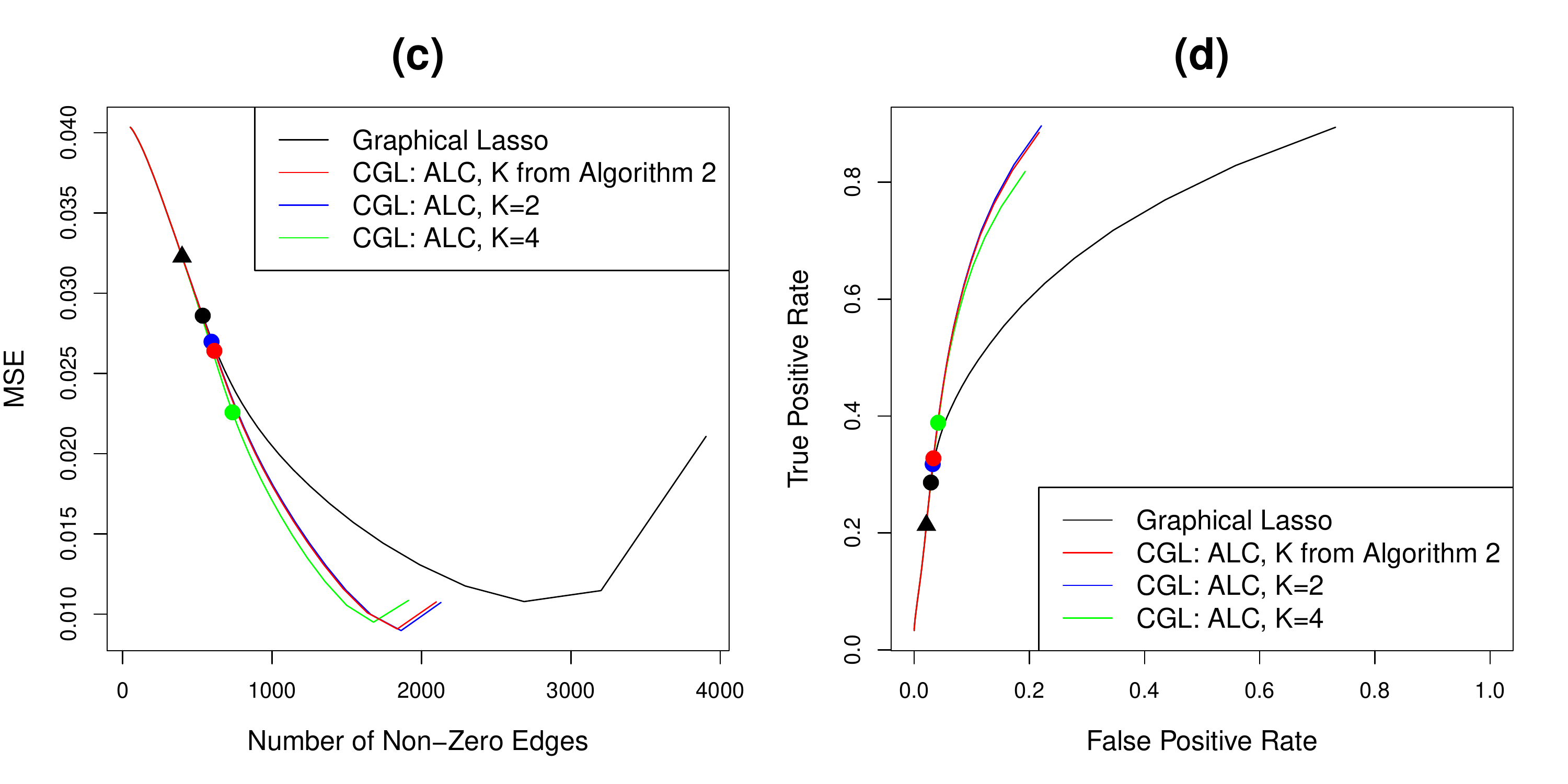}
\end{center}
\caption{Results from graphical lasso, CGL using ALC with various values of $K$ and with $K$ selected automatically using Algorithm \ref{selectK}, averaged over 200 iterations. The true graph is sparse, with two connected components.  (a): Plot of average mean squared error of estimated inverse covariance matrix against the average number of non-zero edges with $n=50$, $p=100$.  (b): ROC curve for the number of edges: true positive rate = (number of correctly estimated non-zero)/(total number of non-zero) and false positive rate =  (number of incorrectly estimated non-zero)/(total number of zero) with $n=50$, $p=100$.  (c)-(d): As in (a)-(b) with $n=200$, $p=100$.
The solid circles indicate $\lambda=\min_k \overline{\lambda}_k - \epsilon$ from Corollary~\ref{chooselambdacorr} and the triangle indicates the choice of $\lambda$ proposed by \protect\citeasnoun{Banerjee}.}%; details are in the text.}
%See text for the explanation of the values represented by the solid circles and triangle.}
\label{n50p100k2}
\end{figure}

Let $\overline{\lambda}_k$ be as defined in  Corollary~\ref{chooselambdacorr}; that is, it is the smallest value of the tuning parameter that will break up the $k$th connected component asymptotically.  The solid circles in Figure~\ref{n50p100k2} correspond to $\lambda= \min_k \overline{\lambda}_k - \epsilon$ for some tiny positive $\epsilon$, i.e., $\lambda$ is  the largest value  such that all $K$ of the connected components are consistently identified according to Corollary~\ref{chooselambdacorr}.  For instance, the black solid circle corresponds to the largest value of $\lambda$ such that the graphical lasso consistently identifies the connected components according to Corollary~\ref{chooselambdacorr}.
From Corollary~\ref{chooselambdacorr}, any value of $\lambda$ to the right of the solid circles should consistently identify the connected components. On the other hand, the black triangle in Figure~\ref{n50p100k2} corresponds to the value of $\lambda$ proposed by \citeasnoun{Banerjee} using $\alpha = 0.05$, as in Equation~\ref{banerjeelambda}.  This choice of $\lambda$ guarantees that the probability of adding edges between two disconnected components is bounded by $\alpha = 0.05$.  
%However, it does not guarantee that the connected components would not be further broken down.

From Figures~\ref{n50p100k2}(a)-(b),  we see that for a given number of non-zero edges, CGL has similar MSE as compared to the graphical lasso, on the region of interest. Also, CGL tends to yield a higher fraction of correctly identified non-zero edges, as compared to graphical lasso.
We see that the value of $\lambda$ proposed by \citeasnoun{Banerjee} leads to a sparser estimate than does Corollary~\ref{chooselambdacorr},   since the black solid triangle is to the left of the black solid circle.
This is consistent with the fact that \citeasnoun{Banerjee}'s choice of $\lambda$ is guaranteed not to erroneously connect two separate components, but is not guaranteed to avoid erroneously disconnecting a connected component.  Moreover, for CGL, the choice of $\lambda$ from Corollary~\ref{chooselambdacorr} results in identifying more true edges, compared to the same choice of $\lambda$ for graphical lasso. This is mainly due to the fact the CGL does better at identifying the connected components. % better identification of connected components in CGL.

In Figures~\ref{n50p100k2}(a)-(b),  $n<p$ and the signal-to-noise ratio is quite low.  Consequently, CGL with $K=4$ clusters outperforms CGL with $K=2$ clusters, even though the true number of connected components is two.  This is due to the fact that in this particular simulation set-up with $n=50$, ALC with $K=2$ has the tendency to produce one cluster containing most of the features and one cluster containing just a couple of features; thus, it may fail to  identify the connected components correctly.
In contrast, ALC with $K=4$ tends to identify the two  connected components almost exactly (though it also creates two additional clusters that contain just a couple of features).

Figures~\ref{n50p100k2}(c)-(d) indicate that when $n=200$ and $p=100$, CGL has a lower MSE than does the graphical lasso for a fixed number of non-zero edges.  In addition, CGL with $K=2$ using ALC has the best performance in terms of  identifying non-zero edges.  The result is not surprising because  when $n=200$ and $p=100$, ALC is able to cluster the features into two clusters almost perfectly.

Overall, CGL leads to more accurate estimation of the inverse covariance matrix than does the graphical lasso when the true covariance matrix is block diagonal. In addition, these results suggest that the method of Algorithm~\ref{selectK} for selecting $K$, the number of clusters, leads to appropriate choices regardless of the sample size.

\subsection{${\bf \Sigma}^{-1}$ approximately block diagonal}
\label{simulation3}
\begin{figure}[htp]
\begin{center}
\includegraphics[scale=0.45]{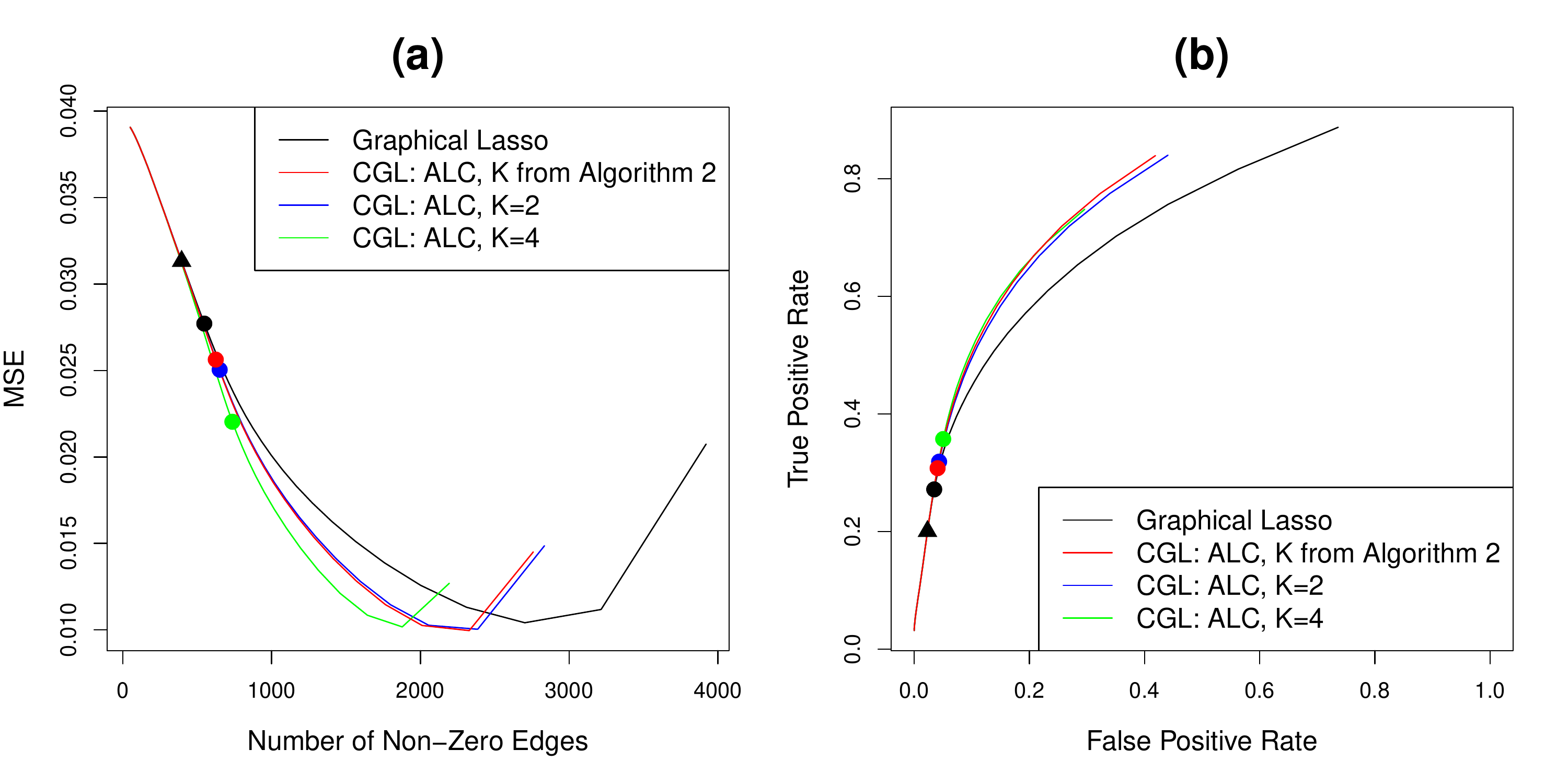}
\includegraphics[scale=0.45]{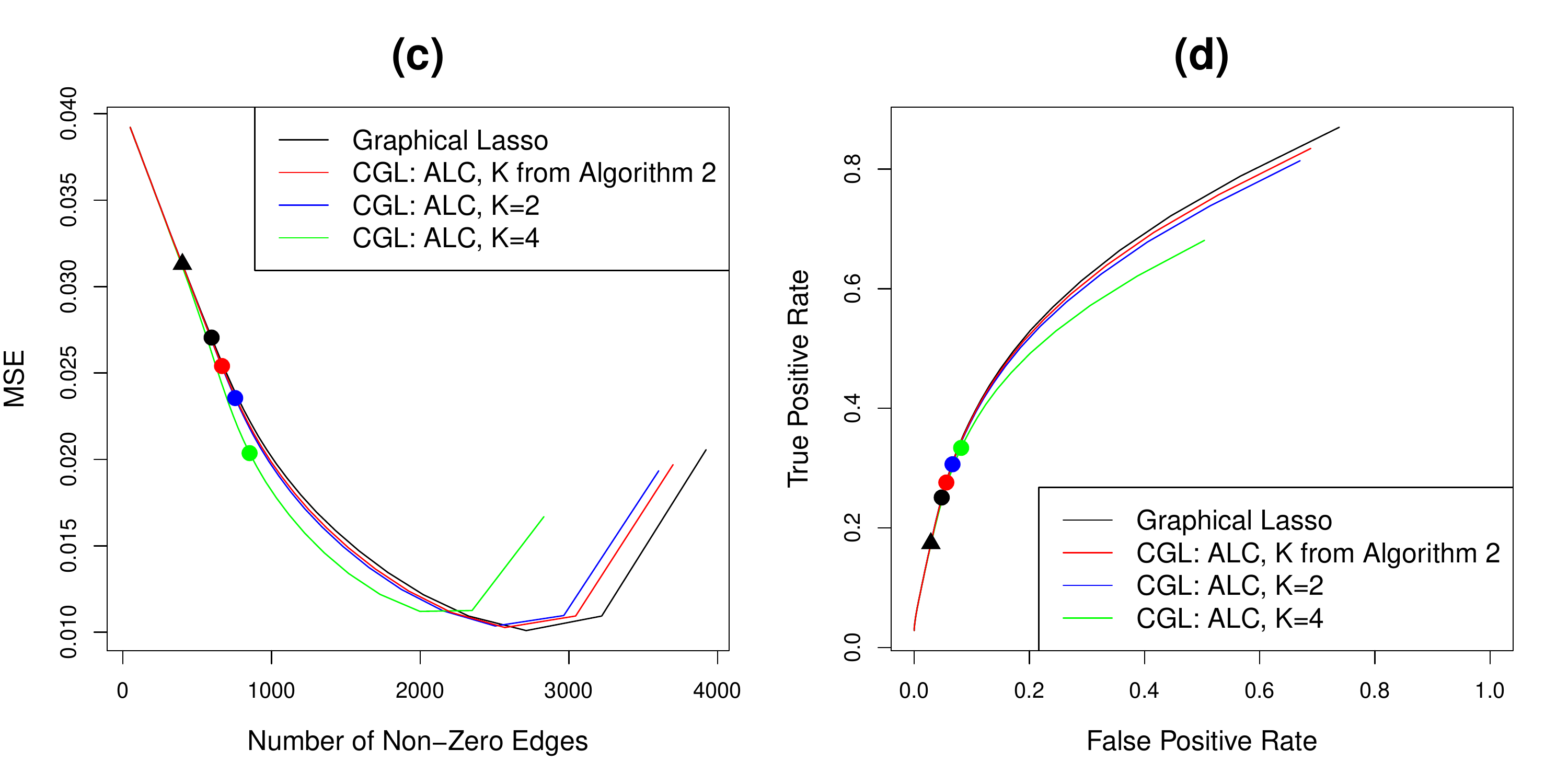}
\caption{As in Figure \ref{n50p100k2}, with $n=200$, $p=100$, and ${\bf \Sigma}^{-1}$ approximately block diagonal.  (a)-(b): $2.5\%$ of the off-block elements do not equal zero. (c)-(d): $10\%$ of the off-block elements do not equal zero.}
\label{abd}
\end{center}
\end{figure}

We repeated the simulation from Section \ref{simulation2}, except that $\mathbf{\Sigma}^{-1}$ is now \emph{approximately} block diagonal.
That is, we generated data according to Algorithm~\ref{precisionmatrix}, but between Steps (c) and (d) we altered the resulting ${\bf \Sigma}^{-1}$ such that
 $2.5\%$ or  $10\%$ of the  elements outside of the blocks are drawn i.i.d. from a $\text{Unif}(-0.5, 0.5)$ distribution.  We considered the case when $n=200, p=100$ in this section.  Results are presented in Figure \ref{abd}.  
 
 From Figures~\ref{abd}(a)-(b), we see that CGL outperforms the graphical lasso when the assumption of block diagonality is only slightly violated.  However, as the assumption is increasingly violated, graphical lasso's performance improves relative to CGL, as is shown in Figures~\ref{abd}(c)-(d).
As in the previous section, we see that Algorithm~\ref{selectK} results in reasonable estimates of the number of clusters.

\section{Application to real data sets}
\label{realdata}
We explore three applications of CGL: to an equities data set in which the features are known to belong to distinct groups, to a webpage data set in which the features are easily interpreted, and to a gene expression data set in which the true conditional dependence among the features is  partially known.  Throughout this section, we choose $\lambda_1= \ldots = \lambda_K =\lambda$ in CGL for simplicity.

\subsection{Equities data}
\label{equities}

We analyze the stock price data from Yahoo! Finance described in \citeasnoun{Hanetal2012}, and available in the \verb=huge= package on \verb=CRAN=  \cite{hugeZhao}.  This data set consists of daily closing prices for stocks in the S\&P 500 index between January 1, 2003 and January 1, 2008.  Stocks that are not consistently included in the S\&P 500 index during this time period are removed.  
This leaves us with  1258 daily closing prices for 452 stocks, which   
%This data set is downloaded from the \verb=R= package \verb=huge=  available on \verb=CRAN=.  
 are categorized into 10 Global Industry Classification Standard (GICS) sectors.
%: Financials (74 stocks), Consumer Discretionary (70 stocks), Information Technology (64 stocks), Industrials (59 stocks), Health Care (46 stocks), Energy (37 stocks), Consumer Staples (35 stocks), Utilities (32 stocks), Materials (29 stocks), and Telecommunications Services (6 stocks).

Let $P_{ij}$ denote the closing price of the $j$th stock on the $i$th day.  Then, we construct a $1257 \times 452$ data   matrix $\mathbf{X}$ whose $(i,j)$ element is defined as $x_{ij} = \log(P_{(i+1)j}/P_{ij})$ for $i=1,\ldots, 1257$ and $j=1, \ldots, 452$.
%Therefore, we have a matrix $\mathbf{X}$.
 Instead of Winsorizing the data as in \citeasnoun{Hanetal2012}, we simply  standardize each stock to have mean zero and standard deviation one.

\begin{figure}[htp]
\centering
\subfigure[]{
\includegraphics[scale=0.45]{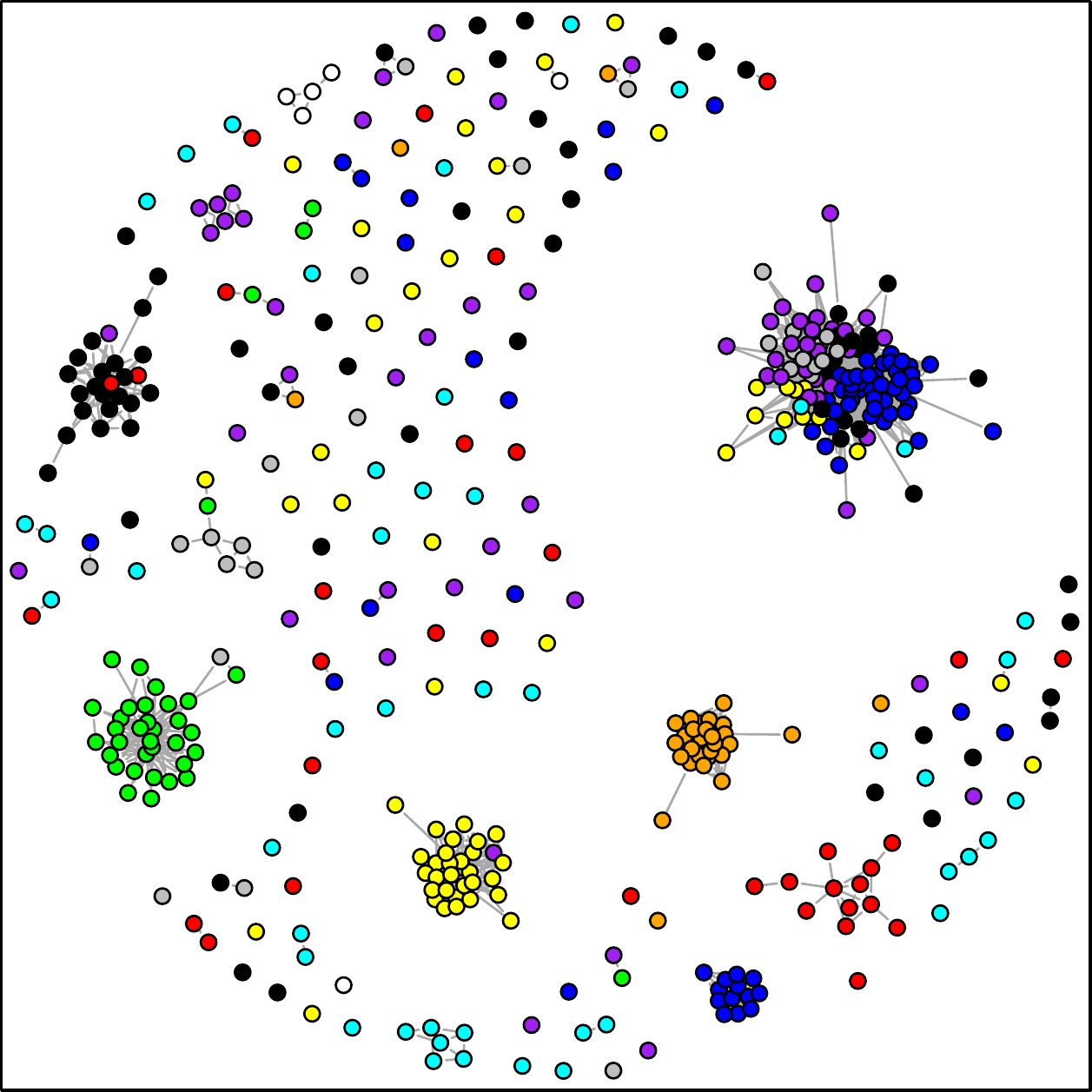}
\label{fig:cgl1}
}
\subfigure[]{
\includegraphics[scale=0.45]{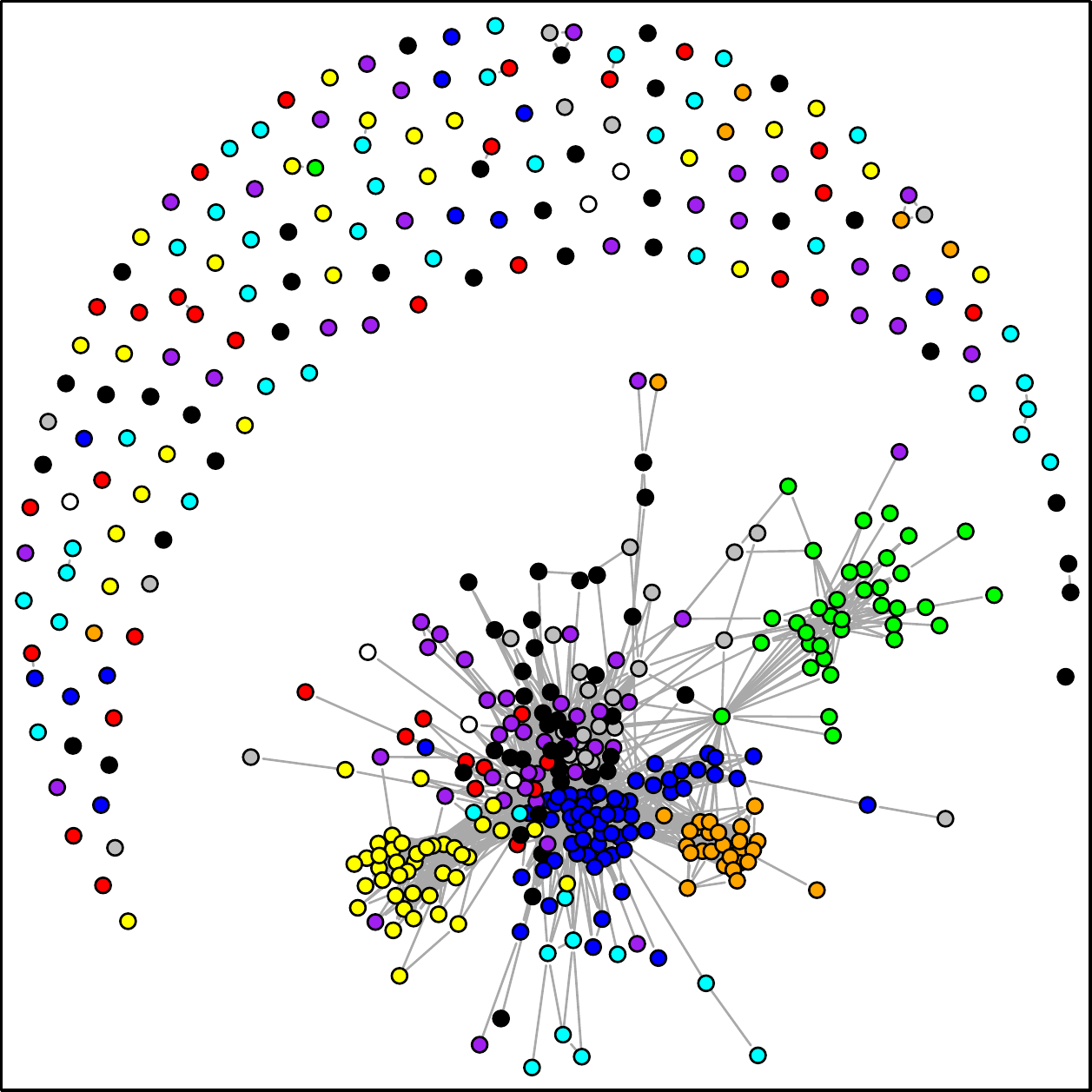}
\label{fig:gl1}
}
\caption{Networks constructed by (a): CGL with $K=10$ and $\lambda = 0.37$ (2123 edges). (b): graphical lasso with $\lambda$ chosen to yield 2123 edges. Each stock was  colored based on its GICS sector: Consumer Discretionary (black), Consumer Staples (red), Energy (green), Financials (blue), Health Care (cyan), Industrials (purple), Information Technology (yellow), Materials (grey), Telecommunications Services (white), and Utilities (orange).}
\label{finance}
\end{figure}

 In this example, the true GICS sector for each stock is known.  However, we did not use this information.  Instead, we performed CGL with CLC and with tuning parameters  $K = 10$ (since there are 10 categories) and $\lambda=0.37$.   The network estimate has  2123 edges.  
We then chose the tuning parameter for the  graphical lasso  in order to obtain the same number of estimated edges. 
 The estimated networks are presented in Figure~\ref{finance}, with nodes colored according to GICS sector.

Figure \ref{finance} reveals that the network estimated by CGL is more easily interpretable than the network estimated by graphical lasso.  For instance, stocks that are categorized as consumer staples (red nodes) are for the most part conditionally independent of stocks in other GICS sectors.  Consumer staples are products such as food, beverages, and household items.  Therefore, stock prices for consumer staples are approximately stable, regardless of the economy and the prices of other stocks.

\subsection{University webpages}
\label{webpage}
In this section, we consider the {university webpages} data set  from the ``World Wide Knowledge Base" project at Carnegie Mellon University.  This data set was preprocessed by \citeasnoun{webpage2011} and previously studied in \citeasnoun{Guo2011}. It includes webpages from four computer science departments at the following universities: Cornell, Texas, Washington, and Wisconsin.
%The webpages were manually classified into seven categories: student, faculty, staff, department, course, project, and other.
 In this analysis, we consider only  student webpages.   This gives us $n=544$ student webpages and $p=4800$ distinct terms that appear on these webpages.

Let $f_{ij}$ be the frequency of the $j$th term in the $i$th webpage.  We construct a $n \times p$ matrix whose $(i,j)$ element is %$\mathbf{X}$ where each element of $\mathbf{X}$ is defined as $x_{ij} =
$\log (1+f_{ij})$.  We selected 100 terms with the largest entropy out of a total of 4800 terms, where the entropy of the $j$th term is defined as $- \sum_{i=1}^n g_{ij} \log (g_{ij})/ \log (n)$ and $g_{ij}=\frac{f_{ij}}{\sum_{i=1}^n f_{ij}}$.  We then standardized each term to have mean zero and standard deviation one.

\begin{figure}[htp]
\centering
\subfigure[]{
\includegraphics[scale=0.45]{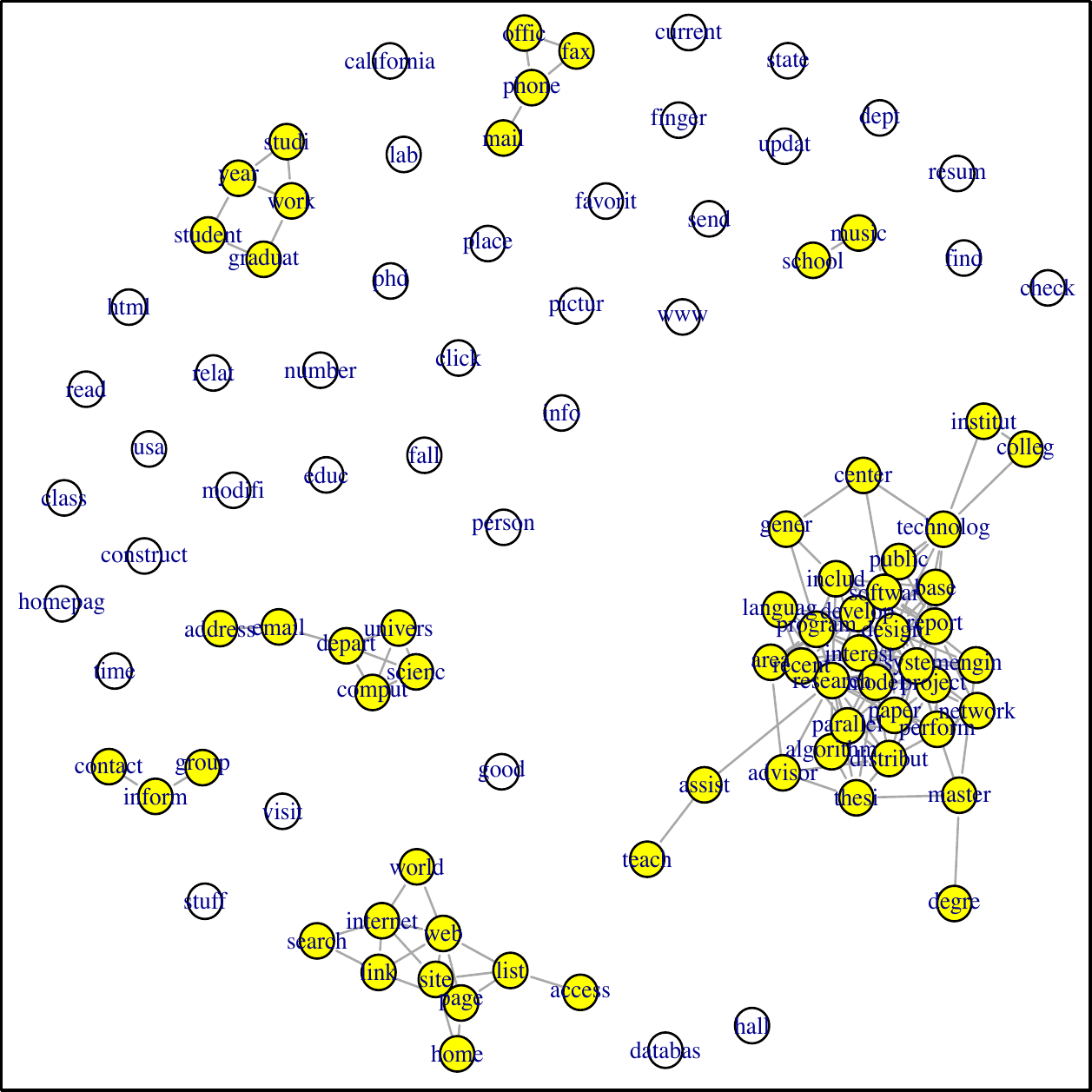}
\label{fig:cgl2}
}
\subfigure[]{
\includegraphics[scale=0.45]{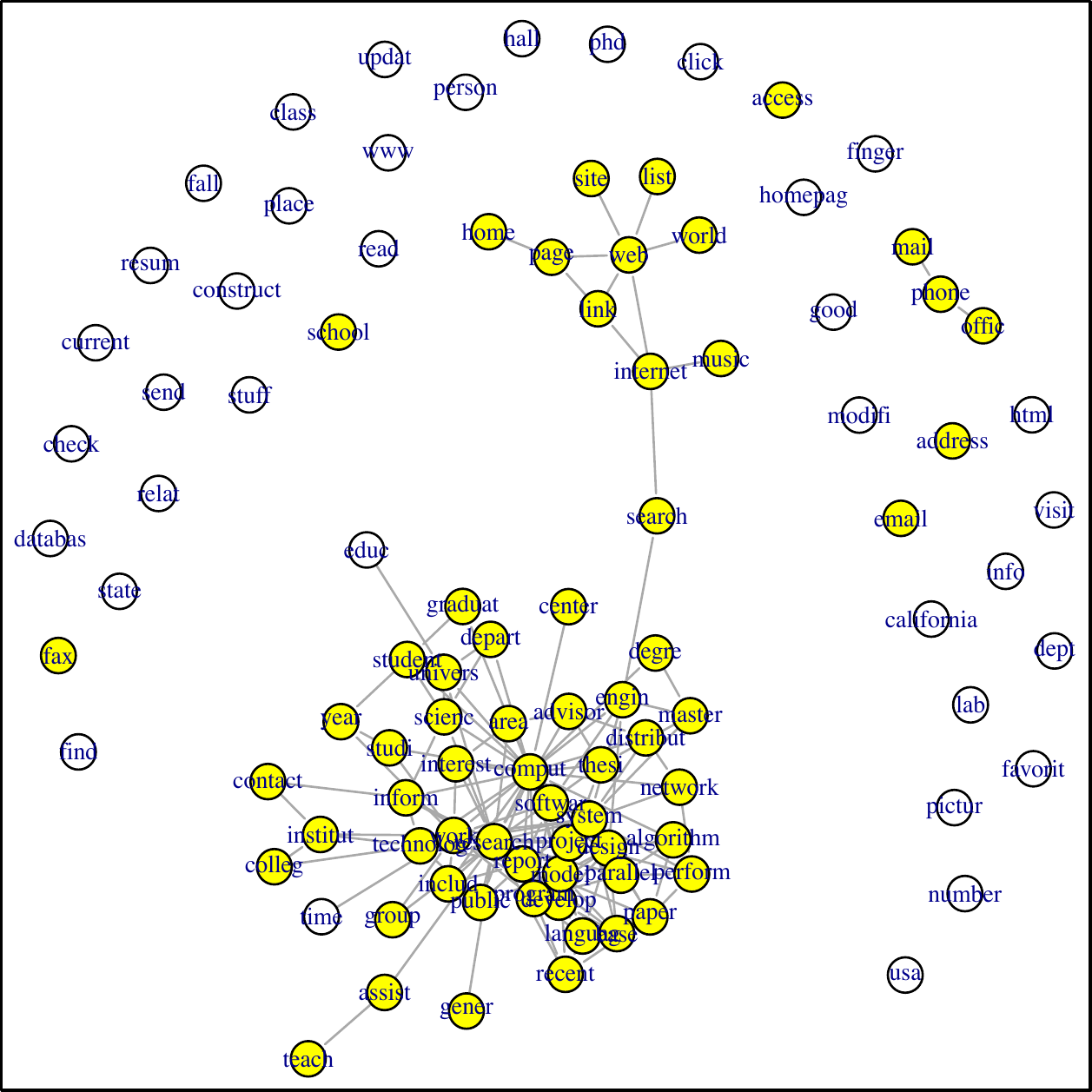}
\label{fig:gl2}
}
\caption{Network constructed by (a): CGL with CLC with $K=5$ and $\lambda = 0.25$ (158 edges). (b): graphical lasso with $\lambda$ chosen to yield 158 edges.  Nodes that are in a multiple-node-subnetwork in the CGL solution are colored in yellow.}
\label{webpagefig}

\end{figure}

For CGL, we used CLC with tuning parameters $K=5$ and $\lambda=0.25$.  The estimated network has a total of 158 edges.
%Note that estimated networks for $K=10$, $K=15$, and also with ALC as the clustering method are  presented in the supplementary materials.
We then performed the graphical lasso with $\lambda$ chosen such that the estimated network has the same number of edges as the network estimated by CGL, i.e., 158 edges.  The resulting networks are presented in Figure~\ref{webpagefig}.  For ease of viewing, we colored yellow all of the nodes in subnetworks that contained more than one node in the CGL network estimate.

%We colored all the connected nodes within each of the subnetwork identified by the cluster graphical lasso.

From Figure \ref{webpagefig}(a), we see that CGL groups related words into subnetworks.  For instance, the terms ``computer'', ``science", ``depart", ``univers" , ``email", and ``address" are connected within a subnetwork.  In addition, the terms ``office", ``fax", ``phone", and ``mail" are connected within a subnetwork.  Other interesting subnetworks include   ``graduate"-``student"-``work"-``year"-``study" and ``school"-``music". In contrast, the graphical lasso in Figure \ref{webpagefig}(b) identifies a large subnetwork that contains so many nodes that interpretation is rendered difficult.  In addition, it fails to identify most of the interesting phrases within subnetworks described above.

\subsection{Arabidopsis thaliana}
\label{Arabidopsis}
We consider the \emph{Arabidopsis thaliana} data set, which consists of gene expression measurements for 118 samples and 39 genes \cite{Rodriguez02}.  This data set has been previously studied by \citeasnoun{WilleEtAl2004}, \citeasnoun{Maetal2007}, and \citeasnoun{HanLiuetal2011}.  It is known that plants contain two isoprenoid biosynthesis pathways, the mevalonate acid (MVA) pathway and the methylerythritol phosphate (MEP) pathway \cite{Rodriguez02}.  Of these 39 genes, 19 correspond to MEP pathway, 15 correspond to MVA pathway, and five encode proteins located in the mitochondrion \cite{WilleEtAl2004,Lange03}.  Our goal was to reconstruct the gene regulatory network of the two isoprenoid biosynthesis pathways.  Although we know that both pathways operate independently under normal conditions, interactions between certain genes in the two pathways have been reported \cite{WilleEtAl2004}.  We expect genes within each of the two pathways to be more connected than genes between the two pathways.

\begin{figure}[htp]
\centering
\subfigure[]{
\includegraphics[scale=0.45]{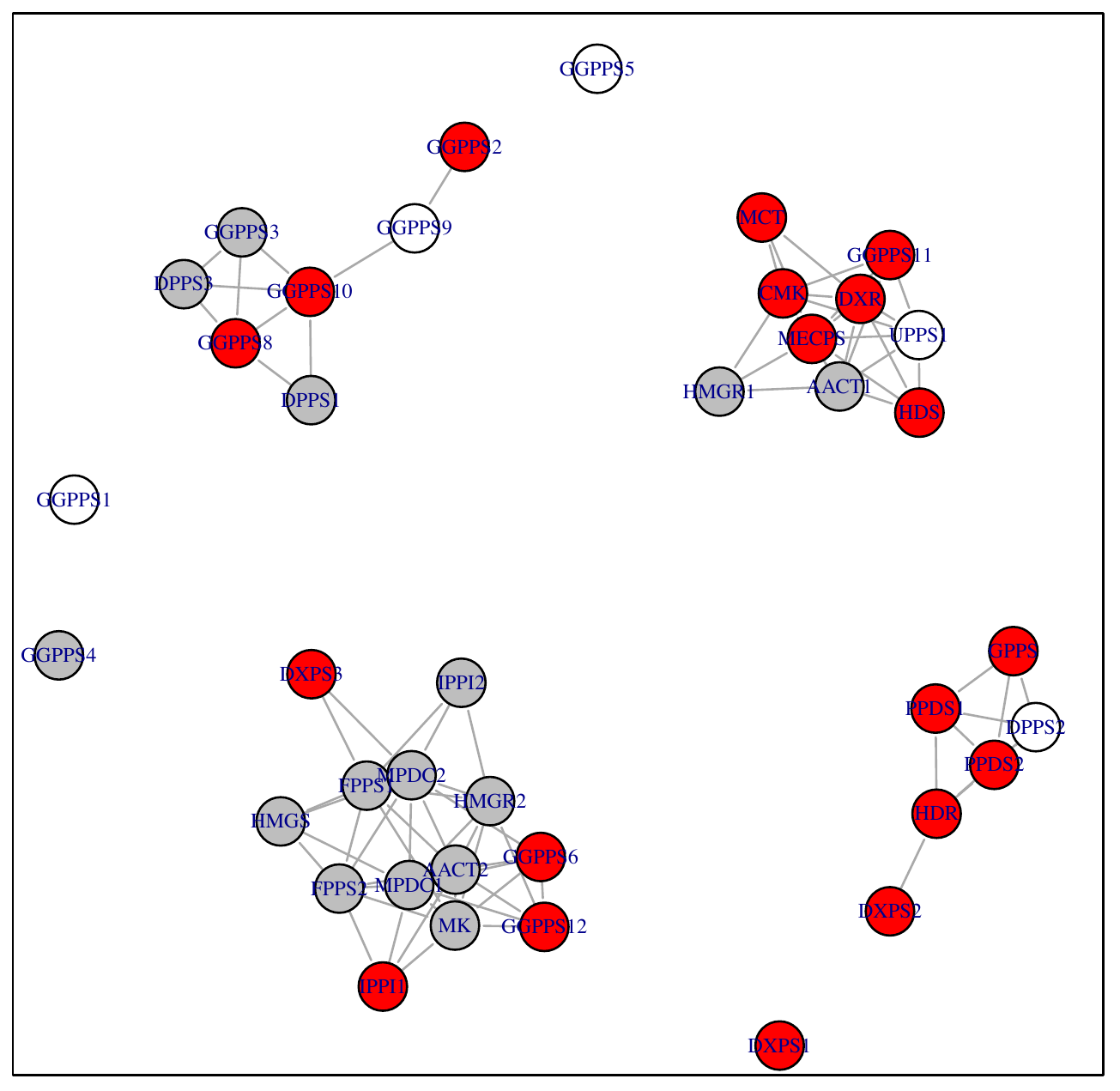}
\label{fig:cgl1}
}
\subfigure[]{
\includegraphics[scale=0.45]{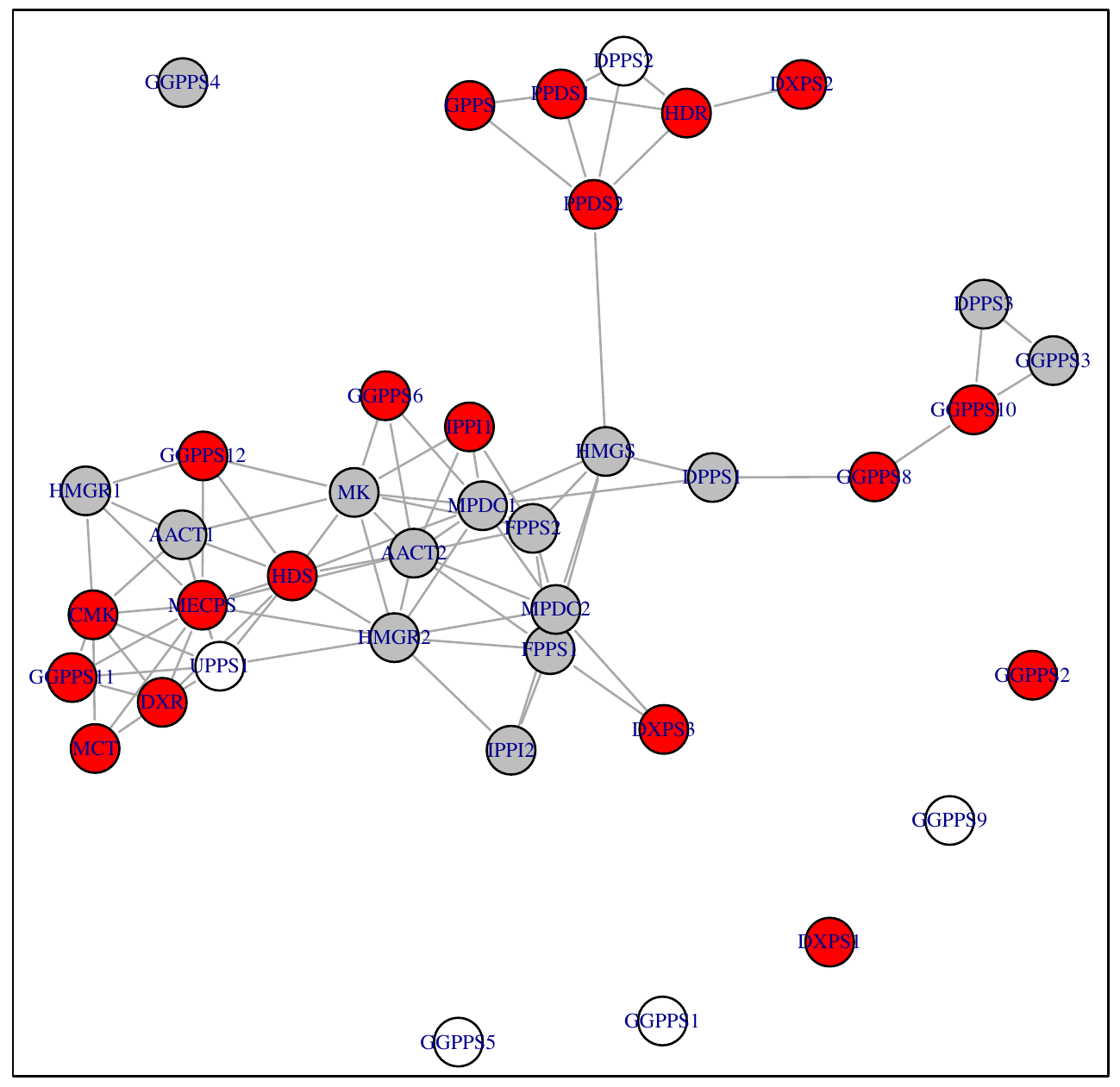}
\label{fig:gl1}
}
\caption{Pathways identified by (a): CGL with $K=7$ and $\lambda = 0.4$ (85 edges), and  (b): graphical lasso with $\lambda$ chosen to yield 85 edges.  Red, grey, and white nodes represent genes in the MEP pathway, MVA pathway, and mitochondrion, respectively. }
\label{cglgl1}
\end{figure}

We began by standardizing each gene to have mean  zero and standard deviation  one.  For CGL, we set the tuning parameters $K=7$ and $\lambda = 0.4$.  The estimated network has a total of 85 edges.  Note that the number of clusters $K=7$ was chosen so that the estimated network has several  connected components that contain multiple genes.  We also performed the graphical lasso with $\lambda$ chosen to yield 85 edges.  The estimated networks are shown in Figure \ref{cglgl1}.  Note that the red nodes, grey nodes, and white nodes in Figure \ref{cglgl1} represents the MEP pathway, MVA pathway, and  the mitochondrion respectively.

From Figure \ref{cglgl1}(a), we see that CGL identifies several separate subnetworks that might be potentially interesting.  In the MEP pathway, the genes DXR, CMK, MCT, MECPS, and GGPPS11 are mostly connected.  In addition, genes AACT1 and HMGR1 which are known to be in the MVA pathway are connected to the genes MECPS, CMK, and DXR.  \citeasnoun{WilleEtAl2004} suggested that AACT1 and HMGR1 form candidates for cross-talk between the MEP and MVA pathway.  For the MVA pathway, genes HMGR2, MK, AACT2, MPDC1, FPPS2, and FPPS1 are closely connected.  In addition, there are edges among these genes and genes IPPI1, GGPPS12, and GGPPS6.  These findings are mostly in agreement with  \citeasnoun{WilleEtAl2004}.   In contrast, the graphical lasso results are hard to interpret, since most nodes are part of a very large connected component, as is shown in Figure \ref{cglgl1}(b).

\section{Discussion}
\label{discussion}

We have shown that identifying the connected components of the graphical lasso solution is equivalent to performing SLC based on $\tilde{\mathbf{S}}$, the absolute value of the empirical covariance matrix.  Based on this connection, we have proposed the cluster graphical lasso, an improved version of the graphical lasso for sparse inverse covariance estimation.

A shortcoming of the graphical lasso is that in order to avoid obtaining a network estimate in which one connected component contains a huge number of nodes, one needs to impose a huge penalty $\lambda$ in (\ref{penlog}).  When such a large value of $\lambda$ is used, the graphical lasso solution  tends to contain many isolated nodes as well as a  number of small subnetworks with just a few edges.   This can lead to an underestimate of the number of  edges in the network.  In contrast, CGL decouples the identification of the connected components in the network estimate and the identification of the edge structure within each connected component.  Hence, it does not suffer from the same problem as the graphical lasso.

In this paper, we have considered the use of hierarchical clustering in the CGL procedure.   We have shown that performing hierarchical clustering on $\tilde{\bf S}$ leads to consistent cluster recovery.  As a byproduct, we suggest a choice of $\lambda_1,\ldots,\lambda_K$  in CGL that yields consistent identification of the connected components.   In addition, we establish the model selection consistency of CGL.  Detailed exploration of different clustering methods in the context of CGL is left to future work.

Equation \ref{penCGL} indicates that CGL can be interpreted as the solution to a  penalized log likelihood  problem, where the penalty function has edge-specific weights that are based upon a previously-obtained clustering of the features. This parallels the adaptive lasso \cite{AdaptiveLasso06}, in which a consistent estimate of the coefficients is used to weight the $\ell_1$ penalty in a regression problem. Other weighting schemes could be explored in the context of (\ref{penCGL}); this is left as a topic for future investigation. 

In this paper, we have investigated the use of clustering before estimating a graphical model using the graphical lasso.
% graphical model estimation using the graphical lasso; 
As we have seen, this approach is quite natural due to a connection between the graphical lasso and single linkage clustering.
However, in principle, one could perform clustering before estimating a graphical model using another technique, such as neighborhood selection \cite{MB2006}, sparse partial correlation estimation
 \cite{Space}, the nonparanormal \cite{LiuLaffertyWasserman2009,XueZou2012}, or constrained $\ell_1$ minimization \cite{CaietalJASA11}. We leave a full investigation of these approaches for future research. 
%Note that the idea of performing clustering on the empirical covariance matrix $\mathbf{S}$ before estimating the inverse covariance matrices is applicable to other procedures as well  \citep{MB2006,Space,LiuLaffertyWasserman2009,CaietalJASA11,XueZou2012}.

\singlespacing
\section*{Acknowledgments} We thank Noah Simon for helpful conversations about the connection between graphical lasso and SLC; Jian Guo and Ji Zhu for providing the university webpage data set in \citeasnoun{Guo2011}; and Han Liu and Tuo Zhao for sharing the equities data set in \citeasnoun{Hanetal2012}.

\section*{Appendix}
\subsection*{Proof of Theorem \ref{main}}
We begin with a lemma \cite{Mirkin1996,Jain1988}.
\begin{lem}
\label{SLC}
Let $C_1, \ldots, C_K$ denote the clusters that result from performing SLC using similarity matrix $ \tilde{\mathbf{S}}$, and cutting the resulting dendrogram at a height of  $0 \le \lambda \le1$.  Also, let $\mathbf{A}$ be a $p \times p$ matrix whose $(j,j')$th element is 1 if $j=j'$ and is $1_{ \{ \tilde{S}_{jj'}  \ge \lambda\} }$ otherwise.  Let $D_1, \ldots, D_R$ denote the connected components of the undirected graph that contains an edge between the $j$th and $j'$th nodes if and only if $A_{jj'} \neq 0$.
Then, $K=R$, and there exists a permutation $\pi$ such that $C_k=D_{\pi(k)}$ for $k=1,\ldots,K$.
\end{lem}
\noindent Combining Theorem \ref{Witten} and Lemma~\ref{SLC}  leads directly to Theorem~\ref{main}.%\subsection*{Proof of Lemma \ref{convergelemma}}

\begin{comment}
\begin{proof}
First, note that by the reverse triangle inequality, $|S_{ij} - \Sigma_{ij}| \ge  |\tilde{S}_{ij} - |\Sigma_{ij}| |$.  This implies that $P[| \tilde{S}_{ij} - |\Sigma_{ij}|   |\ge \delta] \le  P[ |S_{ij} - \Sigma_{ij}| \ge \delta]$.  Then, we have
\begin{equation*}
\begin{split}
P( \max_{i,j} |\tilde{S}_{ij} - |\Sigma_{ij}| |\ge t \sqrt{\frac{\log p }{n}} )
& =  P[  \max_{i,j}  | \tilde{S}_{ij}  - |\Sigma_{ij}| | \ge t \sqrt{\frac{\log p }{n}}] \\
& = P[ \cup_{i,j} ( | \tilde{S}_{ij}  - |\Sigma_{ij}| |  \ge t \sqrt{\frac{\log p }{n}} ) ] \\
& \le  \sum_{i,j} P[| \tilde{S}_{ij} - |\Sigma_{ij}| | \ge t \sqrt{\frac{\log p }{n}}] \qquad \text{by Boole's Inequality}\\
& \le  \sum_{i,j} P[| \tilde{S}_{ij} - \Sigma_{ij}| \ge t \sqrt{\frac{\log p }{n}}] \\
& \le p^2 c_1 \exp (- c_2  t^2 \log p)\\
& = \frac{c_1}{p^{c_2 t^2 - 2}},
\end{split}
\end{equation*}
\noindent where the last inequality follows from Lemma \ref{ravikumarresult}.
\end{proof}
\end{comment}

\subsection*{Proof of Lemma~\ref{clusterconsistent}}
In order to prove Lemma~\ref{clusterconsistent}, we first present an additional lemma.
\begin{lem}
Let $\mathbf{x}_1,\ldots,{\bf x}_n$ be $i.i.d.$ $N(0,\mathbf{\Sigma})$, and assume that $\Sigma_{ii} \le M$ $\forall i$, where $M$ is some constant.  The associated absolute empirical covariance matrix $\tilde{\mathbf{S}}$ satisfies
%the associated empirical covariance matrix $\mathbf{S}= \mathbf{X}^T\mathbf{X}/n$ satisfies
\begin{equation*}
P[  \displaystyle {\max_{i,j}}  | \tilde{S}_{ij}  - |\Sigma_{ij}| |  \ge t \sqrt{\frac{\log p }{n}}] \le \frac{c_1}{p^{c_2 t^2 -2}}.
\end{equation*}
\label{convergelemma}
\end{lem}
%\noindent Proof of Lemma \ref{convergelemma} is provided in the Appendix.  %As a consequence of Lemma \ref{convergelemma}, we note that if $c_2 t^2 > 2$, $p \rightarrow \infty$, and $\frac{\log p}{n} \rightarrow 0$, then $P[  \displaystyle {\max_{i,j}}  | \tilde{S}_{ij}  - |\Sigma_{ij}| |  \ge t \sqrt{\frac{\log p }{n}}] \rightarrow 0$.
In order to prove Lemma~\ref{convergelemma}, first note that
by the reverse triangle inequality, $|S_{ij} - \Sigma_{ij}| \ge  |\tilde{S}_{ij} - |\Sigma_{ij}| |$.  This implies that $P[| \tilde{S}_{ij} - |\Sigma_{ij}|   |\ge \delta] \le  P[ |S_{ij} - \Sigma_{ij}| \ge \delta]$.   Then the result follows from applying
%The proof of Lemma~\ref{convergelemma} follows from applying
Lemma 1 of \citeasnoun{Ravikumar2011}, together with the union bound inequality.

We now proceed with the proof of Lemma~\ref{clusterconsistent}.
\begin{proof}
%By assumption, $\mathbf{\Sigma}$ is block diagonal with $K$ blocks and that $|\Sigma_{ij}| \ge a$ $\forall i,j\in C_k$.  %Let $a$ be the smallest \emph{within-block} element of $|\mathbf{\Sigma}|$: $a = \min_{i,j\in C_k : k=1,\ldots, K} |\Sigma_{ij}|$.
%Then, it is easy to see that t
Let $ a = \min_{i,j\in C_k : k=1,\ldots, K} |\Sigma_{ij}|$.
The assumptions imply that the following holds for performing SLC, ALC, or CLC based on $\tilde{\mathbf{S}}$:
 %if $\max_{i,j} |\tilde{S}_{ij} - |\Sigma_{ij}| | < a/2$, performing hierarchical clustering based on $\tilde{\mathbf{S}}$ correctly identifies the correct clusters.  Conversely,
%\begin{equation*}
%\begin{split}
\[
\{\max_{i,j} |\tilde{S}_{ij} - |\Sigma_{ij}| | < a/2\} \implies \{ \hat{C}_k = C_k \text{ }  \forall k \}
\]
%\{\exists k : \hat{C}_k \ne C_k\} &\implies  \{\max_{i,j} |\tilde{S}_{ij} - |\Sigma_{ij}| | \ge a/2\}.
%\end{split}
%\end{equation*}
Therefore,
\begin{equation*}
\begin{split}
P( \exists k : \hat{C}_k \ne C_k) &\le P[  \displaystyle {\max_{i,j}}  | \tilde{S}_{ij}  - |\Sigma_{ij}| |  \ge a/2]\\
&\le P[  \displaystyle {\max_{i,j}}  | \tilde{S}_{ij}  - |\Sigma_{ij}| |  \ge t \sqrt{\frac{\log p }{n}}]\\
&\le \frac{c_1}{p^{c_2 t^2 -2}},
\end{split}
\end{equation*}
 where the last inequality holds by Lemma \ref{convergelemma}. \end{proof}
 %Allowing both $n$ and $p$ to grow, CGL consistently identifies $C_1,\ldots, C_k$ as $n=\Omega(\log p)$.

\subsection*{Proof of Theorem~\ref{thm:CGLconsistency}}

\begin{proof}
Define $\hat{E}_k$ to be the edge set obtained from applying the graphical lasso to the set of features in $C_k$. Recall that $\hat{C}_k$ is the $k$th cluster estimated in the clustering step of CGL.
We begin by noting that in order for CGL to yield the correct edge set, it must yield the correct edge set within each of the true connected components, and it must also yield no edges between the connected components. In other words,
%\begin{eqnarray}
%P(\hat{E} \neq E) &=& P(\hat{E} \neq E | \exists k: \hat{C}_k \neq C_k) P(\exists k: \hat{C}_k \neq C_k) + P(\hat{E} \neq E |  \nexists k: \hat{C}_k \neq C_k) P(\nexists k: \hat{C}_k \neq C_k ) \nonumber \\
%& \leq & P(\exists k: \hat{C}_k \neq C_k) + P(\hat{E} \neq E |  \nexists k: \hat{C}_k \neq C_k)  \nonumber \\
%& = & P(\exists k: \hat{C}_k \neq C_k) + P(\hat{E} \neq E |   \hat{C}_{k'} = C_{k'} \mbox{ for } k'=1,\ldots,K)  \nonumber \\
%& \leq & P(\exists k: \hat{C}_k \neq C_k) + \sum_{k=1}^K P(\hat{E}_k \neq E_k |   \hat{C}_{k'} = C_{k'} \mbox{ for } k'=1,\ldots,K)  \nonumber \\
%& = & P(\exists k: \hat{C}_k \neq C_k) + \sum_{k=1}^K P(\hat{E}_k \neq E_k |   \hat{C}_{k} = C_{k} )\nonumber.
%\end{eqnarray}
\begin{eqnarray}
%(\hat{E} = E) & = & (\hat{E}_k = E_k \,\forall C_k,\,k = 1, \ldots, K) \cap (\hat{\Theta}_{ij} = 0, (i,j) \notin C_k, \, k = 1, \ldots, K) \nonumber \\
%    & = & (\hat{E}_k = E_k \,\forall C_k,\,k = 1, \ldots, K) \cap (\hat{C}_k = C_k \,\forall \,k = 1, \ldots, K) \nonumber.
(\hat{E} = E) \equiv %& = & (\hat{E}_k = E_k \,\forall \,k = 1, \ldots, K) \cap (\hat{\Theta}_{ij} = 0, (i,j) \notin C_k, \, k = 1, \ldots, K) \nonumber \\
    %& = &
    (\hat{E}_k = E_k \,\forall \,k = 1, \ldots, K) \cap (\hat{C}_k = C_k \,\forall \,k = 1, \ldots, K) \nonumber.
\end{eqnarray}
Thus,
\begin{equation*}
(\hat{E} \neq E) = (\exists k: \hat{E}_k \ne E_k) \cup (\exists k: \hat{C}_k \ne C_k),
\end{equation*}
which implies that
\begin{eqnarray}
P(\hat{E} \neq E) &\leq& P(\exists k: \hat{E}_k \ne E_k) + P(\exists k: \hat{C}_k \ne C_k) \nonumber \\
& \leq & \sum_{k=1}^{K}{P(\hat{E}_k \ne E_k)} + P(\exists k: \hat{C}_k \ne C_k). \nonumber
\end{eqnarray}

By Lemma~\ref{clusterconsistent}, we have that
$P(\exists k : \hat{C}_k \ne C_k) \le \frac{c_1}{p^{c_2t^2-2}}.$   By Theorem 2 of \citeasnoun{Ravikumar2011}, since $n = \Omega( d^2\log(p_{\max}))$ and $\lambda_k = \frac{8}{\alpha} \sqrt{\frac{c_2 (\tau \log p_k + \log 4)}{n}}$, it follows that
\[
    P(\hat{E}_k \ne E_k ) \le \frac{1}{p_{k}^{\tau - 2}} \le \frac{1}{p_{\min}^{\tau - 2}}
\]
%Thus, if $n = \Omega(\tau d^2\log(p_{\max}))$, $P(\hat{E}_k \ne E_k) \le \frac{1}{p_{\min}^{\tau - 2}}$
 for all $k$.  The result follows directly.
 %Hence, by the union bound inequality,
%\[
%P(\exists k : \hat{E}_k \ne E_k ) \le \sum_{k}{P(\hat{E}_k \ne E_k)} \le \frac{K}{p_{\min}^{\tau - 2}}.
%\]
%The result follows from noting that $P(\hat{E} \ne E) \le P(\exists k: \hat{C}_k \ne C_k) + P(\exists k:\hat{E_k} \ne E_k )$.
\end{proof}

\bibliographystyle{agsm}
\bibliography{tibs}

\end{document}